\RequirePackage{fix-cm}
\documentclass[smallcondensed,natbib]{svjour3}     
\smartqed  
\usepackage{graphicx}
\usepackage{mathptmx}      
\usepackage[linesnumbered,ruled,vlined]{algorithm2e}
\usepackage{numprint}
\npthousandsep{\,}
\usepackage{amsmath,amssymb}

\DeclareMathOperator*{\argmax}{arg\,max}

\journalname{Preprint}
\begin{document}

\title{Multi-echelon Supply Chains with Uncertain Seasonal Demands and Lead Times Using Deep Reinforcement Learning
\thanks{This research is supported by the following Brazilian institutions: Fundação de Amparo à Pesquisa do Estado de Minas Gerais (FAPEMIG) and Conselho Nacional de Desenvolvimento Científico e Tecnológico (CNPq).}
}

\titlerunning{Multi-echelon Supply Chains with Uncertain Seasonal Demands and Lead Times Using Deep RL}        

\author{Julio César Alves         \and
        Geraldo Robson Mateus 
}



\institute{J.C. Alves \at
              Department of Applied Computing \\
              Universidade Federal de Lavras \\
              Lavras - MG - Brazil \\
              \email{juliocesar.alves@ufla.br}           
          \and
          G. R. Mateus \at
              Department of Computer Science \\
              Universidade Federal de Minas Gerais \\
              Belo Horizonte - MG - Brazil
}


\date{Received: date / Accepted: date}

\maketitle

\begin{abstract} 
We address the problem of production planning and distribution in multi-echelon supply chains. 
We consider uncertain demands and lead times which makes the problem stochastic and non-linear. 
A Markov Decision Process formulation and a Non-linear Programming model are presented. 
As a sequential decision-making problem, Deep Reinforcement Learning (RL) is a possible solution approach. 
This type of technique has gained a lot of attention from Artificial Intelligence and Optimization communities in recent years. 
Considering the good results obtained with Deep RL approaches in different areas there is a growing interest in applying them in problems from the Operations Research field.
We have used a Deep RL technique, namely Proximal Policy Optimization (PPO2), to solve the problem considering uncertain, regular and seasonal demands and constant or stochastic lead times.
Experiments are carried out in different scenarios to better assess the suitability of the algorithm. 
An agent based on a linearized model is used as a baseline. 
Experimental results indicate that PPO2 is a competitive and adequate tool for this type of problem.
PPO2 agent is better than baseline in all scenarios with stochastic lead times (7.3-11.2\%), regardless of whether demands are seasonal or not.
In scenarios with constant lead times, the PPO2 agent is better when uncertain demands are non-seasonal (2.2-4.7\%).
The results show that the greater the uncertainty of the scenario, the greater the viability of this type of approach.
\keywords{Multi-echelon supply chain \and Stochastic demands \and Stochastic lead times \and Reinforcement learning \and Deep learning \and Proximal Policy Optimization}
\end{abstract}

\section{Introduction}
\label{sec:intro}

\begin{figure*}
  \includegraphics[width=\textwidth]{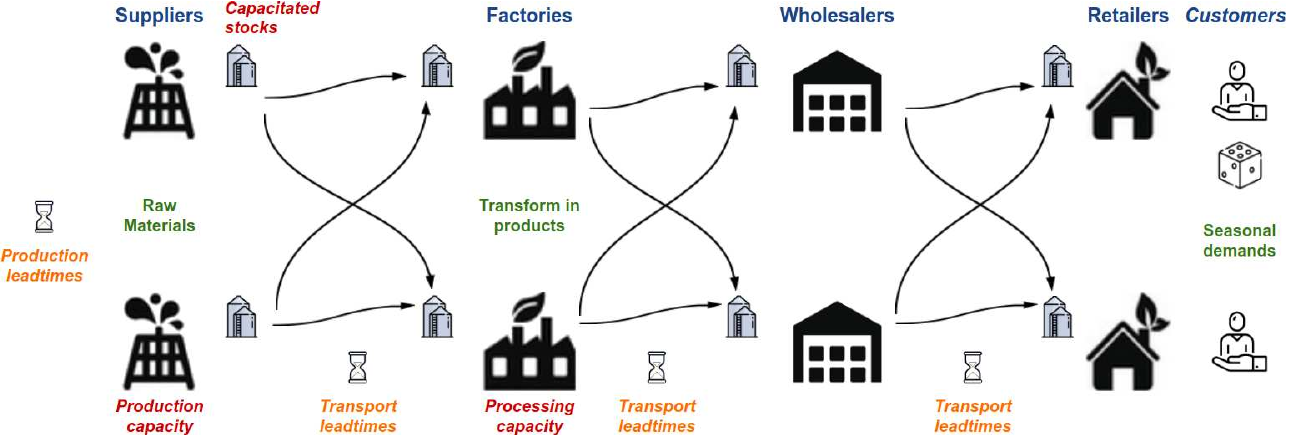}
\caption{Supply Chain addressed: there are four echelons (suppliers, factories, wholesalers, and retailers) with two nodes per echelon. 
All nodes have local capacitated stocks, and suppliers and factories are also capacitated. 
There are stochastic lead times to produce raw material at the suppliers and to transport material from one node to another. 
There are uncertain seasonal customer demands to be met by the retailers.}
\label{fig:supplychain}       
\end{figure*}

The present work uses a Deep Reinforcement Learning (RL) approach to planning the operation of a multi-echelon supply chain with uncertain seasonal demands and lead times.
We consider the case in which the decisions of the whole supply chain are based on ultimate customer demands, and so, there is a central decision-maker and the stages collaborate to minimize total costs.
The supply chain considered is a four-echelon chain composed of two suppliers, two manufacturers (or factories), two wholesalers, and two retailers. 
Suppliers produce and provide raw materials that are processed by manufacturers to generate finished products.
Products are distributed by manufacturers to wholesalers, and wholesalers, in turn, send products to retailers. 
Retailers are responsible for meeting uncertain seasonal customer demands. 
Every node of the chain has a capacitated local stock; suppliers and manufacturers store raw material, while wholesalers and retailers store finished products. 
There are stochastic delays (lead times) to produce raw material at suppliers and to transport material from one node to another. 
There are also maximum capacities regarding production in the suppliers and processing in the factories.
The objective is to operate the entire chain, within a given planning horizon, to meet customer demands and minimize total operating costs. 
Costs are associated with the production and processing of raw materials and with the stock and transport of raw materials and products. 
There is also a penalization cost when customer demand is not met. 
As customers demands and lead times are uncertain, it is not trivial to define the best policy that can meet the customer seasonal demands and, at the same time, minimize the total operating cost. 
Figure \ref{fig:supplychain} illustrates the supply chain scenario addressed, and in Section \ref{subsec:problem_definition}, we present a detailed problem definition.

The problem addressed can be classified as a multi-period production planning and distribution problem under uncertainty in the context of Operations Research (OR). 
Uncertainty in the parameters of a model is very common in real-world OR problems, and usually, it is compensated by safety margins and flexibility buffers; but this generates unused excess capacities and stocks \citep{seelenmeyer2020}. 
Machine Learning (ML) is an alternative approach to solve this issue and has been used to solve problems in many fields. Recent advances, especially with the use of deep neural networks, have leveraged and extended their use. 
RL is a sub-area of ML designed to solve sequential decision-making problems under uncertainty. 
In this context, the problem can be formulated as a Markov Decision Process (MDP), and as the resulting model cannot be solved numerically due to the high dimension of the state space \citep{Laumanns2017}, Deep RL is an appropriate tool to deal with the problem. 
The present work uses a Deep RL approach, namely PPO2\footnote{Earlier works used the term PPO2 to reference the latest version of the algorithm, designed to run in parallel using GPU environments, as is the case of our work. But, recently, the term PPO has been more common in the literature regardless of the version of the algorithm.} (Proximal Policy Optimization), to solve a multi-echelon supply chain problem.
PPO2 was chosen because it achieves high performance in many RL tasks with high-dimensional action spaces \citep{OpenAIBaselinesPPO}.

Some works in the literature use Deep RL on related problems, but they usually deal with smaller supply chain networks, with two-echelon or serial supply chains \citep{Oroojlooyjadid2019,Kemmer2018,Hutse2019,Gijsbrechts2019,Peng2019}.
Considering serial supply chains, the dimensionality of the action space is close to the number of echelons.
But regarding non-serial supply chains, the dimensionality is higher since each node needs to send material to more than one node of the next echelon.
A serial four-echelon supply chain, for instance, has three transport links (supplier-manufacturer, manufacturer-wholesaler, wholesaler-retailer),
while a non-serial four-echelon supply chain with two nodes per echelon has 12 possible transport links.
Therefore, considering the size of the supply chain network used in this work, it is a challenge to solve the presented problem using RL approaches due to the dimensionality of the action space.
\cite{Perez2021} also consider a non-serial four-echelon supply chain, but they do not consider capacitated stocks, seasonal demands, neither stochastic lead times.
According to \cite{morales2020grokking} both Deep RL and OR study decision-making under uncertainty, but problems in the OR field usually have much larger action spaces than those generally treated by Deep RL approaches.
Therefore the use of a Deep RL technique in an OR problem whose action space is continuous and has more dimensions is a contribution of the present research.
Besides that, to the best of our knowledge, this is the first work that handles the problem with stochastic lead times using a Deep RL approach.
In Section \ref{subsec:relatedworks}, we present the related works, classifying them by the type of approach (production planning or order-based) and giving a more detailed explanation about the contributions of the present work.

In previous work \citep{AlvesICCL}, we have used PPO2 to solve the problem considering constant lead times and uncertain regular (nonseasonal) demands in a case study scenario. 
In this paper, we have extended our work to consider uncertain seasonal demands, stochastic lead times, and processing capacities.
We have also deepened the experimental analysis, considering now 17 different scenarios to better assess PPO2 suitability to solve the problem.
The MDP formulation and Non-linear Programming (NLP) model from our previous work were updated to consider uncertain seasonal demands and lead times, and processing capacities.
As the non-linearity of the NLP model comes from the stochastic parameters (demands and lead times), the problem can also be seen as a Linear Program with uncertain parameters.
Therefore, we solve a version of the problem considering forecast (expected) demands and average lead times (a deterministic LP problem).
This solution is encoded in an LP-based agent used as a baseline.
In the experiments, after tuning the hyperparameters of PPO2, several training runs with different random seeds are executed for each scenario.
Studied scenarios consider regular and seasonal demands (with different levels of uncertainty), constant and uncertain lead times, and different stock costs.
The results show that PPO2 can achieve good results in all proposed scenarios.

The remainder of this article is structured as follows.
Section \ref{sec:DeepRL} presents key aspects of Deep RL and the PPO2 algorithm. 
Section \ref{sec:problem} presents the problem definition, related works, and problem modeling (MDP formulation and NLP model).
The methodology is presented in Section \ref{sec:methodology}, including decisions on how to apply PPO2 to solve the problem, the LP-based agent used as a baseline, and the experimental setup. Experimental results are reported and discussed in Section \ref{sec:experiments}. 
A summary of the results and proposed future research directions are finally given in Section \ref{sec:conclusions}.

\section{Deep Reinforcement Learning}
\label{sec:DeepRL}

The concepts presented in this section are mainly based on \cite{sutton2018reinforcement} and \cite{morales2020grokking}.
Deep RL agents can learn to solve sequential decision-making problems under uncertainty formulated as MDPs solely through experience.
The learning process occurs through a trial and error approach.
There is no human labels data, and there is no need to collect or design the collection of data.
Many of the Deep RL techniques are based on an iterative process between the agent and the environment.
At each cycle, the agent observes the environment, takes an action, and receives a reward and a new state (or observation), and the set of these data is called experience.
The purpose of the agent is to maximize the cumulative reward. 
In the case of an episodic task, the idea is to maximize the total reward until the end of the horizon.

The Deep RL agent needs to deal with sequential and evaluative feedback.
Sequential because the action taken in a time step may have delayed consequences.
Evaluative, as the reward is not supervised and, therefore, the agent needs to explore the search space.
The appropriate balance between the gathering of information with the exploitation of current information is known as the exploration vs. exploitation trade-off.
Another important feature of the Deep RL agents is that feedback is sampled.
The reward function is not known by the agent and the state and action spaces are usually large (or even infinite), so the agent needs to generalize from sampled feedback.
The \emph{deep} term in Deep RL refers to the use of artificial neural networks (ANN) with multiple hidden layers.
There are other ways to approximate functions but ANNs often have better performance.
RL is different from Supervised Learning because in Supervised Learning there is a label that specifies the correct action the system should take to a given situation, while in RL, the reward is feedback for the agent but not tell him what would be the correct action.
And RL is not Unsupervised Learning since it is trying to maximize a reward signal instead of trying to find a hidden structure.

Many of the RL techniques are based on an iterative process that alternates between policy evaluation and policy improvement (a pattern also called Generalized Policy Iteration).
The policy evaluation phase calculates the values of a given policy, solving what is called the prediction problem, while policy improvement uses estimates of the current policy to find a new (better) policy.
Alternating between policy evaluation and policy improvement solves the control problem, that is, progressively generates better policies towards optimality.
RL techniques can be based on values or policies. 
In the first case, they learn action (or state) values and use those values to choose a new action.
If based on policies it means that they learn a parameterized policy that allows them to choose actions without the need to consult value estimates.

The basic idea of tabular value-based methods is to use a value function that represents the expected return if the agent follows the current policy $\pi$ from the current state $s$, after taking an action $a$. 
With this approach we can identify the best action to be taken in each state and, during the learning process, to update these values to always improve the policy. 
This kind of approach uses exhaustive feedback meaning that the agents need to have access to all possible samples.
But many problems have high-dimensional state and action spaces (or they could be continuous spaces), and, in such cases, it is not possible to deal with a table for the value function (state or action-values).
The problem is not only with the time and storage constraints, but mainly because many states will probably never be visited during the learning process.
Thus, it is necessary to use methods that deal with sampled feedback and that can generalize from similar states.
One possible approach is to use function approximation to represent the value function instead of tables.
In this approach, when the weights of the function are updated for a state-action pair, the update also impacts the values for other states-action pairs, enabling the desired generalization.
Many of the approximated methods follow the same basic iterative mechanism but using a function approximator like a neural network for instance (e.g, DQN \citep{mnih2013playing} is an approximated approach based on Q-Learning).
But this kind of approach is very limited regarding continuous action spaces because they need to calculate the maximum value over the actions.

Another approach to deal with high-dimensional, and especially continuous, action spaces are policy-based algorithms.
These techniques try to find the best policy directly, instead of learning the value function to derive the best policy.
They have the advantage of being able to learn stochastic policies, and thus exploration is already part of the learned function.
A drawback of policy-based methods is that they can be less sample efficient since it is harder to use off-policy strategies (i.e., it is difficult to reuse a batch of experiences, if it was not generated by the policy being learned).
Policy-gradient methods are a type of policy-based algorithms that solve an optimization problem using the gradient of the performance function of a parameterized policy.
As they follow the gradient concerning stochastic policies, the actions change more smoothly what leads to better convergence properties than value-based methods.
Some of the policy-gradient methods, called actor-critic methods, approximate not only the policy but also the value function.
The actor learns the policy, and the critic learns the value function to evaluate the policy learned by the actor.
In this approach, the value function is used as a baseline and can reduce the variance of the policy gradient objective, and thus often accelerate the entire learning process.

The most powerful actor-critic methods use deep neural networks for both, the actor and the critic, but it is not so easy to obtain good results with these techniques.
In general, they are very parameter sensitive and sample inefficient. 
Proximal Policy Optimization (PPO) was proposed by \cite{schulman2017proximal} to find a balance between implementation, parameterization, and sample complexity (and PPO2 is the latest version, designed to run in parallel using GPU environments).
PPO has similar underlying architecture as previous actor-critic algorithms but innovates with two main contributions.
The first one is a surrogate objective that enables the use of multiple gradient steps on the same mini-batch of experiences.
The second one is the limitation of step size updates.
The goal is to update the policy with a new one that is not so different from the current one.
This has already been proposed in the TRPO method \citep{schulman2017trust}, but while TRPO uses a constrained quadratic objective function (and, thus, it is necessary to calculate second-order derivatives; being hard to parameterize), PPO uses what the authors call a clipped objective function that needs only first-order derivatives; and, at the same time, keeps the sample efficiency and reliable performance of TRPO.
This conservative approach of policy updates prevents performance collapse and enables the reuse of mini-batches of experiences.
Thus the method is more sample efficient and has a lower variance, reaching better performance for many problems.

\begin{figure*}
  \includegraphics[width=\textwidth]{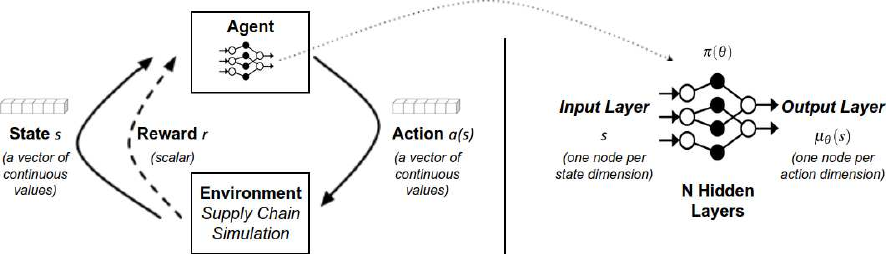}
  \caption{
  Outline of a basic policy-based Deep RL method that learns stochastic policies for problems with continuous action spaces.
  The policy is parameterized by an ANN that receives the state values in the input layer, and outputs mean values for each action value.
  The returned action values are sampled from a Gaussian distribution using such mean values and standard deviations from a vector used to control the exploration of the algorithm.
  }
\label{fig:pg_mechanism}
\end{figure*}

\begin{figure*}
  \includegraphics[width=\textwidth]{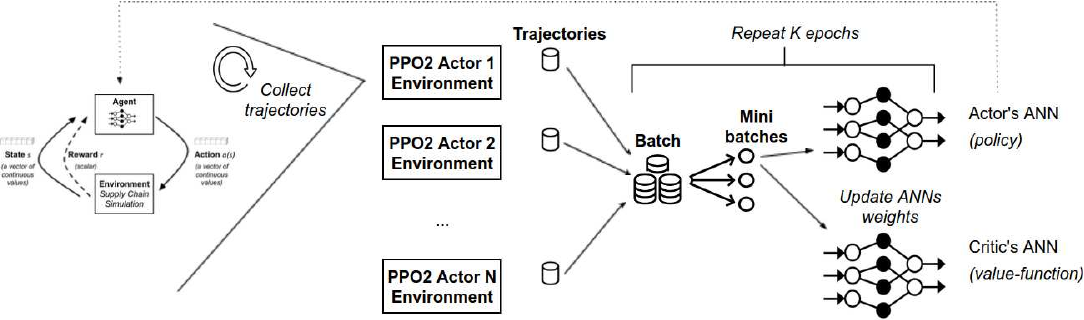}
  \caption{
  PPO algorithm's outline.
  On each step, agent workers collect trajectories using the current policy.
  These data are grouped in a batch of experiences.
  The batch is split into mini-batches that are used to update the actor and critic's ANNs, and this is repeated $k$ times.
  The process is repeated with the updated policy, and this is done until the end of the training.
  }
\label{fig:ppo}
\end{figure*}

Before presenting the PPO algorithm in detail, it is interesting to have a big picture of how policy-based Deep RL methods that learn stochastic policies can handle a problem with continuous action spaces with several dimensions.
Figure~\ref{fig:pg_mechanism} presents the schematic idea of a simple policy-based method.
The learned policy $\pi(\theta)$ is parameterized by a (deep) ANN.
The state $s$ is represented by a vector of continuous values, and ANN's input layer has one node per state value.
The ANN's output layer consists of one node per each action dimension and provides mean values $\mu_{\theta}(s)$ for the actions.
Besides the ANN, the agent has a vector $\sigma_{\theta}$ with standard deviation values for each action value.
The action $a(s)$, returned by the agent, is given by $a(s) = \mu_{\theta}(s) + \sigma_{\theta}(s) \odot z$, where $z \sim \mathcal{N}(0, 1)$ and $\odot$ represents the elementwise product of two vectors \citep{SpinningUp2018}.
Regarding the vector $\sigma_{\theta}$, it can be used to control the exploration of the algorithm.
At the beginning of the training, the vector's values are greater allowing more exploration, and throughout the learning process, the values are slowly decreased to better exploit the agent knowledge.
Another possible approach is to have a separated ANN to learn the standard deviation values.
In this case, the vector is given by $\sigma_{\theta}(s)$, as it depends on the states.
The ANN's weights ($\theta$) can be updated using Stochastic Gradient methods, using the reward received by the agent.
PPO is more complex than this schematic idea as is illustrated in Figure~\ref{fig:ppo}.
First of all, the algorithm uses several workers that collect a bunch of trajectories (experiences) with the current policy and groups them in a batch.
For $k$ epochs, this batch is randomly split into mini-batches, that, in turn, are used to update the weights of the actor and critic's ANNs.
After the update, the process is repeated using the new policy.
These steps are executed until a stopping criterion is reached.

To present the algorithm in more detail, the PPO's pseudo-code is shown in Algorithm \ref{alg:PPO}.
In line 1 policy parameters, $\theta$, and value-function parameters, $\phi$, are initialized (randomly, e.g.). 
The for-loop initialized in line 2 refers to the total amount of time steps the algorithm will be running and, so, it is a parameter to be defined for the problem.
The for-loop shown in lines 3-5 is responsible for collecting the buffer of experiences (trajectories).
Each actor (potentially in parallel) runs the current policy for $\tau$ time steps and collects a set of experiences (line 4), then, in line 5 the advantage function is calculated for each experience. 
PPO uses a truncated version of generalized advantage estimation (GAE), given by $\hat{A}_t = \delta_t + (\gamma \lambda)\delta_{t+1} + ... + (\gamma \lambda)^{\tau-t+1}\delta_{\tau-1}$, where $t$ is the time step in range $[0,\tau]$; $\delta_t = r_t + \gamma V(s_{t+1}) - V(s_t)$; $V$ is the value-function; and $\gamma$ and $\lambda$ are parameters of the algorithm.
The advantage function indicates how much better it is to take one action instead of following the current policy, so indicates the advantage of choosing an action instead of the default one. 
The buffer of experiences collected by all actors is joined in a batch in line 6.
Lines 7-11 perform multiple gradient steps with the collected experiences.
The batch of experiences is randomly split in mini-batches (line 8) and each mini-batch of experiences is used to update ANN weights for both policy (line 10) and value-function (line 11).
This process of randomly splitting the buffer and updating ANNs is repeated $K$ times.

\begin{algorithm}
\DontPrintSemicolon
Initialize $\theta$ and $\phi$ (policy and value-function parameters, respectively)\;
\For{i=0,1,2,...}{
    \For{actor=1,2,...,N} {
        Run policy $\pi(\theta)$ for $\tau$ time steps and collect the trajectories\;
        Calculate advantage estimates $\hat{A}_1,...,\hat{A}_\tau$\ based
        on value-function $V_{\phi}$\;
    }
    Form a $batch$, of size $N\tau$, with collected trajectories and advantages\;
    \For{k=1,2,...,K} {
        Shuffle $batch$ and split into $minibatches$\;
        \ForEach{$minibatch$}{
            Update the policy by maximizing the objective $\theta_{k+1} = \underset{\theta}{\argmax} \underset{s,a \sim \pi_{\theta_k}}{{\mathrm E}}\left[ L(s,a,\theta_k, \theta)\right]$ via stochastic gradient ascent\;
            Update $\phi$ fitting value function by regression on mean-squared error via gradient descent\;
        }
    }
}
\caption{{\sc Proximal Policy Optimization (PPO)}}
\label{alg:PPO}
\end{algorithm}

The objective function $L$ used to update the policy parameters (line 10) is presented in Equation \ref{eq:PPO_obj_function}.

\begin{equation} \label{eq:PPO_obj_function}
    L(\theta) = L^{CLIP}_t(\theta) - c_1 L^{VF}_t(\theta) + c_2 S[\pi_{\theta}](s_t),
\end{equation}

where $L^{CLIP}$ is given by Equation \ref{eq:PPO_Lclip}, $L^{VF}_t$ is the value-function loss, $S$ is an entropy bonus to assure enough exploration, and $c_1$ and $c_2$ are parameters of the algorithm. 
Regarding $L^{CLIP}$ function\footnote{$clip(x, min, max)$ gives $x$ for $min \leq x \leq max$, $min$ for $x<min$ and $max$ for $x>max$.}: $\pi_{\theta}$ and $\pi_{\theta_k}$ are the new and current policies, respectively; and $\epsilon$ is a parameter which indicates how far away the new policy is allowed to deviate from the current one \citep{SpinningUp2018}.

\begin{equation} \label{eq:PPO_Lclip}
    L^{CLIP}(\theta) = min\Big(\frac{\pi_{\theta}(a|s)}{\pi_{\theta_k}(a|s)}\hat{A}_t, clip\big(\frac{\pi_{\theta}(a|s)}{\pi_{\theta_k}(a|s)},1-\epsilon,1+\epsilon\big)\hat{A}_t\Big)
\end{equation}

Besides the parameters presented, the number and size of the ANN's hidden layers and the size of the update step (learning rate) used in the gradient methods are other parameters of PPO.
As mentioned by \cite{henderson2019deep}, there are implementation details that affect PPO performance that is not fully presented by \cite{schulman2017proximal}.
In Section \ref{sec:experiments}, the values used for all parameters and the implementation version of the algorithm are presented.

\section{The Problem}
\label{sec:problem}

This section is organized as follows.
Problem definition is presented in Section \ref{subsec:problem_definition}, and related works are presented in Section \ref{subsec:relatedworks}. 
MDP formulation and NLP model are presented in sections \ref{subsec:mdp} and \ref{subsec:nlp}, respectively.

\subsection{Problem Definition}
\label{subsec:problem_definition}

In this section, we intend to formalize the supply chain problem we are considering in this work.
Related works that apply RL techniques to multi-echelon supply chains are usually inspired by the so-called beer distribution game \citep{Sterman1989}.
This game was proposed to study the bullwhip effect by analyzing the ordering behavior of individuals with only local information in a four-echelon linear supply chain.
In this context, each node of the chain can be viewed as an independent actor that needs to attend to demands from its direct successor in the supply chain and to decide how much to buy (to order) from its predecessor.
Regarding the beer game example, there is a supplier, a factory, a wholesaler, and a retailer.
The retailer attends final customer orders and needs to place orders to the wholesaler. 
The wholesaler, in turn, attends the orders from the retailers and places orders to the factories, and so on. 
The supplier (first-echelon) obtains material from an external source.
Therefore, in this case, which we call an order-based approach, the decisions go somehow upstream the chain since an order of a node is for the precedent echelon.
Another possible approach is to consider the decisions of the whole supply chain based on ultimate customer demands, as recommended by \cite{Lee1997} to counteract the bullwhip effect.
In this context, there is a single agent, a central decision-maker that controls all the chain operations, and the problem can be seen as a multiperiod production planning problem \citep{pinedo2009planning,Stadtler2015}.
In this setting, the decisions in each time step are: how much raw material to produce in each supplier, how much material to transport from a node to its successors, and how much material to store in each node's local stock.
Therefore the decisions to be taken go somehow downstream the chain, as the transport of material is decided from each node to the subsequent echelon's nodes.
In this work, we consider the second case, i.e., a multiperiod production planning problem with a single agent.

A supply chain can operate with backlog or lost sales approaches.
In case of backlog, a client demand in a time step can be met later, while with the lost sales approach an unattended client demand is discarded.
We consider a lost sales approach in this work, and a penalization cost is incurred when a client demand is not attended.
Another important definition is that we consider the case of continuous manufacturing (process) industries \citep{pinedo2009planning}, so all the quantities of materials are continuous values.
Nevertheless, we believe the methodology used here could be adapted for discrete industries, and suggestions on how to do this are given in Section \ref{subsec:applying_ppo2}.

We present now the dynamics of the supply chain we study in this work.
All scenarios experimented consider a four-echelon supply chain with two nodes per echelon.
At the beginning of the planning horizon (time step $t=0$), there is an initial amount of material (raw material or product) stored on each node's local stock, raw materials are being produced by the suppliers (that it will become available on the next time steps), and raw materials and products are being transported from each node to its successors in the next echelon.
There are also customer demands for each retailer that will need to be attended at the next time step ($t=1$).
The steps presented below are repeated until the end of the planning horizon:
\begin{enumerate}
    \item The central decision-maker (the agent) decides the amount of material to be produced on each supplier, and the amount of material to be transported from each node to each of its successors. 
    The amount of material to be kept in stock is indirectly defined by the remaining material in each node.
    Retailers are not controlled since they attend to customer demands whenever possible.
    \item A new time step is considered (now $t=t+1$).
    \item Material flow:
    \begin{itemize}
        \item Raw materials that were being produced in each supplier and that are now available, due to expired lead times, are stored in the supplier's stock.
        \item Raw materials and products in transport, with expired lead times, are delivered in each node and stored in its stock.
        \item Possible excess of materials (beyond stock's capacity) is discarded and penalization costs are incurred.
    \end{itemize}
    \item Retailers attend to customer demand using material from their stocks. If there is not enough material, penalization cost is considered for each unit of the missing product.
    \item Agent's decisions are followed:
    \begin{itemize}
        \item Production of raw material is triggered in each supplier. The lead time of production is realized, i.e., it is defined when the raw material will be available in the supplier's stock.
        \item For each node (except retailers) the amount of material to be transported to each of its successors is removed from stock and shipped to the corresponding successor. 
        The lead time of transport is realized,  i.e., it is defined when the material will be available in the successor's stock.
        In the case of factories, raw materials are processed into finished products before the shipment.
    \end{itemize}
    \item Uncertain (potentially seasonal) customer demands of each retailer for the next time step are realized.
\end{enumerate}

It is important to highlight that the demands are uncertain, and in each time step ($t$) the agent only sees the realization of the demand for the next time step ($t+1$).
As the retailers are not controlled by the agent, the agent cannot take any advantage of knowing only the demands of the next time step.
Another point is that, as the lead times for producing raw material and transporting material are uncertain, the agent needs to make such decisions without knowing exactly when the materials will be available (or delivered).
The lead time values are realized after the decisions have been made, and only in the next state, the agent will know when the material will arrive.
Furthermore, as it will be presented in the MDP formulation (Section \ref{subsec:mdp}), the agent only knows the exact amount of material that will arrive (or be delivered) in the next time step. 
The quantities of material arriving in the time steps after the next one are added together into a single value.

Regarding solution methods, there are different approaches depending on how to handle uncertain parameters like demands and lead times.
In practice, it is common to use demand forecasts and average lead times and, in this case, the problem becomes deterministic and can be solved by LP models \citep{Stadtler2015}.
Another approach is to solve the problem taking into account the uncertainty of the demands and lead times by using methods that can handle stochasticity.
In this case, the problem is non-linear and can be solved with techniques like Stochastic Programming or Deep RL.
We use a Deep RL approach to solve the problem and an agent based on an LP model solution as a baseline.
There is also the option to handle the problem using single-agent or multi-agent approaches.
Multi-agent approaches are more common when the problem is modeled considering that each node of the chain is an independent actor.
In our approach, the problem is solved with a single agent that takes several different decisions at the same time.

\subsection{Related Works}
\label{subsec:relatedworks}

There are some works that use RL for supply chain operation problems, but many of them are based on tabular RL techniques, like Q-Learning, for example \citep{Giannoccaro2002,Chaharsooghi2008,Mortazavi2015}.
As tabular techniques cannot properly handle problems like that addressed in this work, with continuous state and action spaces, we focus here on the most recent works that use Deep RL approaches to solve similar problems.
Firstly, we present works that deal with the problem in a production planning approach, as we do in this paper.
Then we present related works that deal with order-based approaches.

\cite{Kemmer2018} use Approximate SARSA and three versions of Vanilla Policy Gradient (VPG, also called REINFORCE) on a two-echelon supply chain. 
The scenario consists of a factory and one to three warehouses with increasing demands, no lead times, and a horizon of 24 time steps.
The state is composed of the stock levels and the demands of the last two time steps; the actions refer to the factory production and product transportation (but the action space is reduced to only 3 production and transportation levels); 
and the rewards are the profit, considering operating costs and backlogs.
($r$-$Q$)-policy, a minimum stock approach, is used as a baseline. 
All agents are better than baseline on the scenario with only one warehouse, and two versions of VPG improve over the baseline in the scenario with 3 warehouses. 
The work is extended by \cite{Hutse2019}, including deterministic (non-zero) lead times, two product types, continuous action spaces, and four types of stochastic demand scenarios.
The author uses a DQN (Deep Q-Network) for discrete actions and DDPG (Deep Deterministic Policy Gradient) for continuous actions. 
The state is composed of the stock, production, transport and the last $x$ (a parameter) demands, for each node-product combination. 
The actions are, for each product, how much to produce and to send for each retailer (using aggregated levels in the discrete case and limiting the maximum action values in the continuous case). 
The rewards are the profit, considering operating costs (including stock-outs). 
The baseline is the ($r$-$Q$)-policy and the agents are better than the baseline in all scenarios (1 factory, 2 or 3 retailers, and 1 or 2 products).
\cite{Peng2019} use VPG in a capacitated supply chain with one factory warehouse and three retailers (with balanced and unbalanced costs), regular and seasonal stochastic demands, and constant lead times.
The state is composed of the stock levels and the last two demands, and the actions are how much to produce and to send to each retailer. 
As the actions are state-dependent, they use two mechanisms to treat the inherent difficulty of using this approach with neural network outputs. 
The rewards are the profit, considering operating costs and penalization by not satisfying demand. 
The Deep RL agent achieves better results than baseline, ($r$-$Q$)-policy, in all experimented scenarios.

We now present related works that handle the problem using an order-based approach.
\cite{Gijsbrechts2019} propose a proof of concept by using Deep-RL on three different problems: dual-sourcing or dual-mode, lost sales, and multi-echelon inventory models. 
In the multi-echelon setup, demands are uncertain and regular, and lead times are deterministic. 
States are represented by the stock levels and orders of the warehouses and retailers, while the actions are the orders from each node (aggregated by state-dependent base-stock levels). 
They apply A3C (Asynchronous Advantage Actor-Critic) on two different scenarios with one warehouse and ten retailers, considering stochastic demands. 
The A3C agent performs better than a base-stock policy used as a baseline. 
\cite{Oroojlooyjadid2019} uses a customized DQN to solve the MIT Beer Game, a four-echelon linear supply chain, considering deterministic and stochastic demands, and deterministic lead times. 
The author treats the problem as decentralized, with multi-cooperative agents. 
Each agent only knows the local information, and to avoid competition, there is an engineered mechanism to provide feedback to each agent at the end of an episode.
In the experiments, only one agent uses DQN and the others follow a base-stock heuristic. 
The state is composed of the stock levels, demands, arriving orders, and arriving products, from the past $m$ (a parameter) time steps. 
The actions refer to how much more or less to order than the received order, and the used intervals are $[-2,2]$ and $[-8,8]$. 
The rewards are the stock plus backlog costs. 
Experiments show that using DQN for one node achieves better results than using the base-stock policy for all nodes.
\cite{Hachaichi2020} use PPO and DDPG to solve an inventory replenishment problem in a two-echelon supply chain.
There is one distribution center and three stores, with local capacitated stocks.
Supply is unlimited, customer demands are nonseasonal (and lower than stock's store capacities), lead times are constant, and a planning horizon of 52 time steps is considered.
States are composed of stock levels, material in transport, and customer demands from the past $m$ (a parameter) time steps.
Actions are represented by the orders from the distribution centers and stores.
The objective is to maximize profit and, so, rewards are sales minus stock and order costs.
Experiments with a fixed scenario show that DDPG results are unstable and PPO achieves a 6.4\% gap from a baseline where all observed demands are satisfied.
\cite{Geevers2020} uses PPO in three problem cases, considering linear, divergent, and two-echelon supply chains.
Considering the last scenario, an industrial case study, stocks are capacitated and supply is unlimited.
The problem is solved considering one type of product, constant lead times, uncertain (nonseasonal) demands, and a planning horizon of 50 time steps.
States are composed of total stock, total backorders, the stock levels, and, for each pair of nodes, the backorders, and material in transport;
and actions are represented by the order quantity for every stock point.
The objective is to minimize the total holding and backorder costs.
In the experiments, the PPO agent achieves results with a large variance.
Considering 10 training runs, some runs are better than the base stock baseline while others yield poor results.
The author argues that the adopted method is unstable and, therefore, it is not yet fitted to be used in practice.
\cite{Perez2021} use PPO to solve an inventory management problem in a make-to-order four-echelon supply chain. 
Nodes have local stocks (without capacity limits), production is limited, there is a single product and a planning horizon of 30 time steps is considered.
There is one retailer to attend to uncertain (nonseasonal) demands, and the lead times are heterogeneous (without uncertainty).
States are composed of the demand, the stock levels, and the material in transport, and the actions are represented by the reorders quantities.
The rewards are the profit calculated for each node of the chain.
They experiment with a case study scenario considering backlogging and lost sales options, and compare PPO with four LP models (deterministic and multi-stage stochastic, considering rolling or shrinking horizon). 
All LP models are better than PPO when backlogging is considered.
In the lost sales scenario, PPO is only better than one of the models (deterministic LP with rolling horizon).
The authors argue that the PPO solution has a more balanced load and it could potentially have greater resilience to disruptions.

Table \ref{tab:rel_works} shows a comparison of related works and our approaches.
We have grouped the works by type of approach (production planning or  order-based). 
For each work is presented the supply chain configuration (nodes per echelon, products, and planning horizon), the type of uncertainty for demands (regular or seasonal), indication if lead times are deterministic or stochastic, information about state and action spaces (if they are continuous or discrete, and the number of dimensions), and Deep RL technique and baseline used.
As can be seen on the table, to the best of our knowledge, this paper, and our previous work, are the first ones that deal with the problem considering a production planning approach in a supply chain with more than two echelons and including stochastic lead times.
As we consider more than one node per echelon, the number of dimensions of the state and action spaces are larger than similar works, and, therefore, the problem is bigger and harder to solve.
When considering also papers with order-based approaches, most of the related works deal with two-echelon supply chains, with constant lead times and smaller action spaces.
Like our work, \cite{Perez2021} use PPO and LP-based baseline in a four-echelon supply chain.
However, they handle the problem in an order-based approach and do not consider capacitated stocks, seasonal demands, neither stochastic lead times, the state and action spaces are discrete, and the planning horizon is smaller.
Moreover, their experimental methodology lacks important steps such as hyperparameter tuning, training runs with different seeds, and rewards normalization \citep{henderson2019deep,stable-baselines3}, while all these steps are considered in the present paper.
\cite{Geevers2020} has also used PPO and considered continuous action space with many dimensions, but he uses regular demands and deterministic lead times, in a two-echelon supply chain, and has achieved unstable results.
Our main contributions to the literature, considering the best of our knowledge, can be summarized as follows.
\begin{enumerate}
    \item The present and our previous works are the first ones to use Deep RL to handle the problem with a production planning approach in a supply chain with more than two echelons.
    \item This is the first work to use a Deep RL method to solve the problem with stochastic lead times (even considering works with order-based approaches).
    \item This work and \cite{Perez2021} are the only ones that deal with a four-echelon non-serial supply chain using Deep RL. 
    This leads to a problem with more state and action space dimensions and, therefore, harder to solve.
    But, different from \cite{Perez2021}, we consider a production planning approach with seasonal demands, stochastic lead times, capacitated stocks, continuous state and action spaces, and a larger planning horizon.
    \item Finally, we have conducted a robust experimental methodology, achieving good results with PPO2 considering continuous action space with more dimensions than related works.
\end{enumerate}

\begin{table}
\caption{Comparison with related works.
The works are grouped by type of approach: production planning, or order-based. 
For chain configuration: $Config.$ means the number of nodes per echelon, $P$ the number of products, and $H$ the planning horizon.
Column $Dem$ indicates if demands are regular or seasonal; column $Lt$ shows if lead times are deterministic or stochastic.
In columns \textit{States} and \textit{Actions}, $D (M) $ or $C (M)$ indicate if it is a discrete or continuous $M$-dimensional space.
\textit{RL Alg.} column presents the RL technique used, and the last column shows the baseline.
The values refer to the most complex experimented scenarios of each work.}
\label{tab:rel_works}
\addtolength{\tabcolsep}{-3pt}
\begin{tabular}{llllllllll}
\hline\noalign{\smallskip}
\textbf{Authors} & \multicolumn{3}{c}{\textbf{Chain}} & \textbf{Dem}  & \textbf{Lt} & \textbf{States} & \textbf{Actions} & \textbf{RL Alg.} & \textbf{Baseline}  \\
  & Config. & P & H & & & & & \\
\noalign{\smallskip}\hline\noalign{\smallskip}
\multicolumn{4}{l}{\textit{\quad Production planning approaches}} \\
\noalign{\smallskip}
\cite{Kemmer2018} & 1-3 & 1 & 24 & S & Det  & D (10) & D (4) & VPG & $(r,Q)$ \\ 
\cite{Hutse2019} & 1-3 & 2 & 52 & S & Det & D (30) & C (4) & DQN, DDPG & $(r,Q)$ \\
\cite{Peng2019}  & 1-3 &   1     & 25 & S & Det & C (12) & C (4) & VPG & $(r,Q)$ \\
\cite{AlvesICCL} & 2-2-2-2 & 1 & 360 & R & Det & C (27) & C (14) & PPO2 & LP-based \\
\textbf{This work} & 2-2-2-2 & 1 & 360 & S & Sto & C (27) & C (14) & PPO2 & LP-based \\
\noalign{\smallskip}\hline\noalign{\smallskip}
\multicolumn{4}{l}{\textit{\quad Order-based approaches}} \\
\noalign{\smallskip}
\cite{Gijsbrechts2019} & 1-10 & 1 & \textit{cont.} & R & Det & D (35) & D (2) & A3C & base stock \\
\cite{Oroojlooyjadid2019} & 1-1-1-1 & 1 & \textit{fixed} & R & Det & D (50) & D (1) & cust. DQN & base stock \\
\cite{Hachaichi2020} & 1-3 & 1 & 52 & R & Det & C (48) & C (4) & PPO, DDPG & \textit{custom} \\
\cite{Geevers2020}   &   4-5   & 1 &  50 & R & Det & C (48)   &  C (9)  & PPO & base stock \\
\cite{Perez2021} & 2-3-2-1 & 1 &  30 & R & Det & D (68) & D (11) & PPO & LP-based \\
\noalign{\smallskip}\hline
\end{tabular}
\end{table}

\subsection{MDP Formulation}
\label{subsec:mdp}

Modeling a complex sequential decision-making problem as an MDP is one of the most important tasks to solve the problem using RL techniques.
Formulating an MDP means defining the states, actions, rewards, and environment's dynamics (or transition function) to be used to solve the problem.
The formulation presented here is an extension of our previous work \citep{AlvesICCL} to include uncertain seasonal demands and lead times, and processing capacities.

\subsubsection{State Space}

\begin{figure*}
  \includegraphics[width=\textwidth]{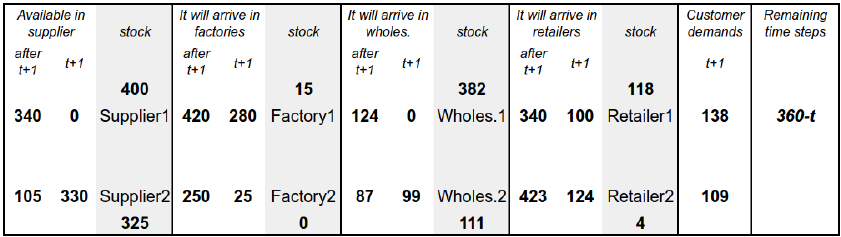}
\caption{Example of a state for a given time step $t$.}
\label{fig:mdp_state}
\end{figure*}

A state for a given time step $t$ is a 27-dimensional continuous vector with the following values (an example of a state is presented in Figure~\ref{fig:mdp_state}).
\begin{itemize}
    \item The current stock level of each node.
    \item For each supplier: 
    \begin{itemize}
        \item the amount of raw material being produced and that will be available in the next time step ($t+1$);
        \item the sum of the amount of raw material being produced and that will be available in the time steps after the next one.\footnotemark
    \end{itemize}
    \item For each other node:
    \begin{itemize}
        \item the amount of material in transport that will arrive in the node in the next time step (it is the sum of material sent by the two nodes from the predecessor echelon);
        \item the sum of the amount of material in transport that will arrive in the time steps after the next one.\footnotemark[\value{footnote}]
    \end{itemize}
    \item The final customer demands of each retailer for the next time step ($t+1$).
    \item The number of remaining time steps until the end of the episode.
\end{itemize}

\footnotetext{The idea of using a summarized value for material available after the next time step is to avoid increasing state space with information that is not so precise, since with stochastic lead times these values are more likely to change on the next time steps.
}

In order to obtain good results with Deep RL algorithms like PPO2, it is important to normalize the state values \citep{stable-baselines3}.
In Section~\ref{subsec:applying_ppo2} we present how the state values normalization is done in our experiments.

\subsubsection{Action space}

\begin{figure*}
  \includegraphics[width=0.8\textwidth]{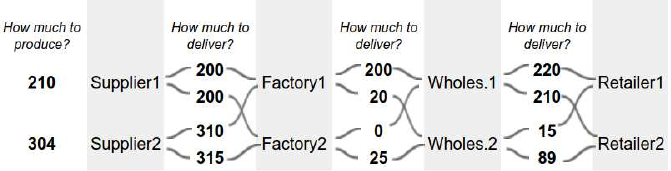}
\caption{Example of an action: decisions are regarding production of raw materials in the suppliers and transportation between nodes (stock levels are indirectly defined and retailers are not controlled).}
\label{fig:mdp_action}
\end{figure*}

An action is a 14-dimensional continuous vector with the following values (an example of an action is presented in Figure~\ref{fig:mdp_action}).
\begin{itemize}
    \item The amount of material to produce in each supplier.
    \item For each node (except retailers).
    \begin{itemize}
        \item The amount of material to deliver to each of the two nodes of the next echelon.
    \end{itemize}
\end{itemize}

As mentioned in the problem definition, the amount of material to be kept in stock is indirectly defined by the remaining material in each node, and retailers are not controlled since they attend to customer demands whenever possible.

One important practical aspect of defining the action representation is how to handle possible unfeasible actions.
Regarding the production of material in each supplier, it is simple to avoid unfeasible values by limiting the action to the supplier's capacity.
But, for material to be transported from one node to another the situation is more complex since the node's stock level changes throughout the simulation.
In Section~\ref{subsec:applying_ppo2} we present how we deal with this challenge and generate only feasible actions in our experiments. In the same section, we present how we handle action values normalization since it is important to obtain good results with Deep RL algorithms like PPO2 \citep{stable-baselines3}.

\subsubsection{Environment's dynamics}

As we use a model-free RL approach, a simulation of the supply chain is used to represent the environment’s dynamics. 
Almost all supply chain operations are simulated in a deterministic way, that is, any amount of material defined by the action values (to be supplied or transported) is followed by the simulation, as the actions always represent feasible quantities.
The non-deterministic behavior of the simulation is due to uncertain customer demands and lead times.
Customer demands and lead times are sampled in each time step from a statistical distribution and are realized as presented in the problem definition (Section \ref{subsec:problem_definition}).
There is also a treatment for the excess of material in stocks, which can happen because a node can receive more material than it can store since we need to sum the amount of material arriving from different nodes and its current stock level.
The simulation considers that all arriving material needs to pass by the stock, even if it is not kept for the next time step.
The excess of material is discarded and a related penalization cost is incurred.

\subsubsection{Rewards} 

The design of the rewards is crucial for the success of an RL algorithm. 
For many problems, it is not clear the best way to define the reward to the agent, especially because in many cases it can be easy to define feedback at the end of an episode (success or fail), but it can be difficult to define feedback on each simulation step. 
However, in the proposed supply chain problem, as it is a cost minimization problem, seems to make a lot of sense to use the negative of the total operating costs as the reward\footnote{We have also experimented with the inverse of the operating costs multiplied by a constant as the reward, but the PPO2 algorithm could not learn with this approach.}.
Therefore the reward used is the negative of the sum of all incurred costs at a time step (production, transportation, manufacturing, stocks, and penalization by material discarded due to stock capacities and by not meet customer demands).

\subsection{Non-Linear Programming Model}
\label{subsec:nlp}

In this section, an NLP model of the problem is presented.
The model is an extension of our previous work \citep{AlvesICCL} to include uncertain seasonal demands and lead times, and processing capacities.
The model is nonlinear only because demands and lead times parameters are stochastic. 
If we consider forecast values for demands and average values for lead times the model becomes an LP model.

The sets of the model are presented in Table \ref{tab:nlp_sets}, where $q$ is the number of nodes of the chain, $h$ is the planning horizon (episode length) and $l^{max}$ is the maximum possible lead time value.
Time step $0$ refers to the initial state of the chain.
Variables of the model are presented in Table \ref{tab:nlp_variables} and the parameters in Table \ref{tab:nlp_param}.
Note that $p_{ijn}$, $f_{n}$, and $t_{ijnm}$ are binary, and all other parameters are integers.
$f_{n}$ is $1$ for all factories and $0$ for the other nodes.
$p_{ijn}$ is used to define which nodes are suppliers and also to map the lead times.
There is one stochastic lead time value for each supplier on each time step, so that $p_{ijn}=1$ only if $n$ is a supplier and the lead time realization to produce material on that supplier at time step $i-j$ is $j$, otherwise $p_{ijn}$ is zero.
Similarly, $t_{ijnm}$ defines which node pairs have transport links and also maps the lead times.

\begin{table}
\caption{Sets of the NLP model}
\label{tab:nlp_sets}
\begin{tabular}{ll}
\hline\noalign{\smallskip}
\textbf{Set} & \textbf{Description} \\
\noalign{\smallskip}\hline\noalign{\smallskip}
\(N = \{1,...,q\}\) & set of the supply chain nodes \\
\(I = \{0,1,...,h+l^{max}\}\) & set of time steps (or periods) \\
\(J = \{1,...,l^{max}\}\}\) & set of possible lead times \\
\noalign{\smallskip}\hline
\end{tabular}
\end{table}

\begin{table}
\caption{Variables of the NLP model}
\label{tab:nlp_variables}
\addtolength{\tabcolsep}{-2pt}
\begin{tabular}{ll}
\hline\noalign{\smallskip}
\textbf{Variable} & \textbf{Description} \\
\noalign{\smallskip}\hline\noalign{\smallskip}
$S_{in}$   & the stock level of node $n$ on time step $i$ \\
$T_{ijnm}$ & amount of material sent from node $n$ at time step $i-j$ to arrive on node $m$ on time step $i$ \\
$P_{ijn}$  & amount of raw material produced by a supplier $n$ on time step $i-j$ to be available on time step $i$ \\
$F_{in}$   & amount of raw material processed by a factory $n$ on time step $i$ \\
$D^e_{in}$ & excess of material discarded for exceeding the stock capacity of node $n$ on time step $i$ \\
$D^d_{in}$ & amount of missing products to meet customer demand by a retailer $n$ in time step $i$ \\
\noalign{\smallskip}\hline
\end{tabular}
\end{table}

\begin{table}
\caption{Parameters of the NLP model}
\label{tab:nlp_param}
\addtolength{\tabcolsep}{-2pt}
\begin{tabular}{ll}
\hline\noalign{\smallskip}
\textbf{Par.} & \textbf{Description} \\
\noalign{\smallskip}\hline\noalign{\smallskip}
$q$       & number of nodes in the chain \\
$h$       & planning horizon (episode length) \\
\noalign{\smallskip}\hline\noalign{\smallskip}
\(c_n^s\) & cost of stocking one unit of material at node \(n\) \\
\(c_n^p\) & cost of producing one unit of raw material at node \(n\) \\
\(c^f_n\) & cost of processing one unit of raw material at node \(n\) \\
\(c^t\)   & cost of sending one unit of material from one node to another \\
\(c^e\)   & cost of one unit of material discarded by exceeding stock capacity \\
\(c^d \)  & the cost incurred by unmet demand (for each unit of product) \\
\noalign{\smallskip}\hline\noalign{\smallskip}
\(b^s_n\) & the stock capacity of the node \(n\) \\
\(b^p_n\) & production capacity of the supplier \(n\) \\
\(b^f_n\) & processing capacity of the node \(n\) \\
\(b^t_n\) & the maximum amount of material that can be sent from node \(n\) \\
 \noalign{\smallskip}\hline\noalign{\smallskip}
\(r_n\)      & processing ratio at node \(n\) \\
\(l^{max}\)  & maximum lead time (for production in the suppliers and transport)  \\
\(l^{avg}\)  & average lead time (for production in the suppliers and transport) \\
\(f_{n} \)   & indicate if it is possible to process raw material on node \(n\) \\
\(p_{ijn}\)  & indicate if it is possible to produce raw material on node \(n\) at time step \(i-j\) to be available at time step \(i\) \\
\(t_{ijnm}\) & indicate if it is possible to send material from node \(n\) at time step \(i-j\) to arrive on node \(m\) at time step \(i\) \\
\noalign{\smallskip}\hline\noalign{\smallskip}
\(s_{n}\)    & initial stock level on node \(n\) \\
\(p_{in}\)   & initial amount of material produced by supplier \(n\) that will be available on time step \(i\) \tiny{(defined for $i \leq l^{avg}$)} \\
\(t_{inm}\)  & initial amount of material sent by node \(n\) to node \(m\) that will be available on time step \(i\) \tiny{
(defined for $i \leq l^{avg}$)} \\
\noalign{\smallskip}\hline\noalign{\smallskip}
\(d_{in}\)   & stochastic customer demand to be met by node \(n\) on time step \(i\) \\
\noalign{\smallskip}\hline
\end{tabular}
\end{table}

The target of the NLP model is to minimize the total operating cost and the objective function\footnote{Regarding results presented in Section \ref{sec:experiments}, the costs related to initial materials (stocks, supplied, and transport) are not considered in the objective function.} is given by Equation \ref{eq:nlp_obj_fun}.

\begin{equation} \label{eq:nlp_obj_fun}
    min \quad \sum_{i \in I} \sum_{n \in N}\Big( c^s_n S_{in} + c^f_n F_{in} + c^e D^e_{in} + c^d D^d_{in} \Big) + \sum_{i \in I} \sum_{j \in J} \sum_{n \in N}\Big(c^p_n P_{ijn} + \sum_{m \in N} c^t T_{ijnm} \Big)
\end{equation}
    
The constraints are defined as follows.
Constraints \ref{eq:stock_const} control the stock, transport of material, and demands.
Capacities are handled by constraints \ref{eq:supcap_const}, \ref{eq:proccap_const}, \ref{eq:trancap_const}, and \ref{eq:stocap_const}.
Constraints \ref{eq:proc_const} are used to calculate the amount of raw material processed at factories.
And, finally, constraints \ref{eq:init_sup_const}, \ref{eq:init_tran_const}, and \ref{eq:init_sto_const} are used to take into account the initial supplied, transported, and stocked materials.

\begin{equation} \label{eq:stock_const}
\begin{split}
S_{in} = S_{(i-1)n}+ \sum_{j \in J} P_{ijn} + \sum_{j \in J} \sum_{m \in N} T_{ijmn} - D^e_{in}  - r_n \Big(\sum_{j \in J} \sum_{m \in N} T_{(i+j)jnm}\Big) - d_{in} + D^d_{in} \\ \forall \quad i \in \{1,...,h\}, n \in N
\end{split}
\end{equation}

\begin{equation} \label{eq:supcap_const}
 0 \leq P_{ijn} \leq p_{ijn} b^p_n \quad \forall \quad i \in I, j \in J, n \in N 
\end{equation}
\begin{equation} \label{eq:proccap_const}
0 \leq F_{in} \leq b^f_n \quad \forall \quad i \in I, j \in J, n \in N
\end{equation}
\begin{equation} \label{eq:trancap_const}
 0 \leq T_{ijnm} \leq t_{ijnm} b^t_n \quad \forall \quad i \in I, j \in J, n \in N, m \in N
\end{equation}
\begin{equation} \label{eq:stocap_const}
 0 \leq S_{(i-1)n} + \sum_{j \in J} P_{ijn} + \sum_{j \in J} \sum_{m \in N} T_{ijmn} - D^e_{in} \leq b^s_n  \quad \forall \quad i \in \{1,...,h\}, n \in N
\end{equation}

\begin{equation} \label{eq:proc_const}
F_{in} = f_n r_n \Big(\sum_{j \in J} \sum_{m \in N} T_{(i+j)jnm}\Big) \quad \forall \quad i \in \{1,...,h\}, n \in N
\end{equation}
\begin{equation} \label{eq:init_sup_const}
 P_{iin} = p_{in} \quad \forall \quad i \in \{1,...,l^{avg}\}, n \in N
\end{equation}
\begin{equation} \label{eq:init_tran_const}
 T_{iinm} = t_{inm} \quad \forall \quad i \in \{1,...,l^{avg}\}, n \in N, m \in N
\end{equation}
\begin{equation} \label{eq:init_sto_const}
S_{0n} = s_{n} \quad \forall \quad n \in N
\end{equation}

\section{Methodology}
\label{sec:methodology}

In this section, we describe the methodology used to solve the supply chain problem with uncertain seasonal demands and lead times.
In Section~\ref{subsec:scenarios}, we present the 17 scenarios we have used in the experiments to evaluate the suitability of the PPO2 algorithm to solve the problem.
We present how we have applied the algorithm in Section~\ref{subsec:applying_ppo2}.
The main objective is to present the normalization of state and action values we have used to obtain better results with the method.
In Section~\ref{subsec:lpagent}, we describe how the LP agent is built and used as a baseline in the experiments.
In Section~\ref{subsec:lower_bounds}, we present how we use the LP model with perfect information to calculate lower bounds for each experiment.
Finally, the three-phase experimental methodology (hyperparameter tuning, training, and evaluation) is presented in Section~\ref{subsec:exp_meth}.

\subsection{Experimental Scenarios}
\label{subsec:scenarios}

We have considered several scenarios to assess the suitability of the PPO2 algorithm to solve the proposed problem.
The parameters that are common to all scenarios are presented in Table~\ref{tab:scenarios_params}, following the notation from Section~\ref{subsec:nlp}.
Chain configuration and costs are the same used by \cite{AlvesICCL}, and all costs are applied by a unit of raw material or product. 
Initial values and capacities were defined according to the range of demand values.

\begin{table}
\caption{Parameters common to all scenarios }
\label{tab:scenarios_params}
\addtolength{\tabcolsep}{-2pt}
\begin{center}
\resizebox{\textwidth}{!}{%
    \begin{tabular}{llll}
    \hline\noalign{\smallskip}
    \textbf{Group} & \textbf{Param.} & \textbf{Value} & \textbf{Details} \\
    \noalign{\smallskip}\hline\noalign{\smallskip}
    Chain &  q        & 8 & 2 suppliers, 2 factories, 2 wholesalers, and 2 retailers \\
          & \(f_{n}\) & 1 & for factories (0 for the other nodes) \\
          & \(r_n\)   & 3 & processing ratio for factories (1 for the other nodes) \\
    \noalign{\smallskip}\hline\noalign{\smallskip}
    Horizon      &  h      & 360 & episode length \\
    \noalign{\smallskip}\hline\noalign{\smallskip}
    Costs & \(c_n^s\) & 1     & stock costs for all nodes \\
          & \(c_n^p\) & 6,4   & production cost for each supplier, respectively \\
          & \(c^f_n\) & 12,10 & processing cost for each factory, respectively  \\
          & \(c^t\)   & 2     & transport cost on the whole chain \\
    \noalign{\smallskip}\hline\noalign{\smallskip}
    Pen. costs & \(c^e\) & 10  & cost of material discarded by exceed stock capacity \\
               & \(c^d\) & 216 & cost incurred by unmet demand \\
    \noalign{\smallskip}\hline\noalign{\smallskip}
    Capacities & \(b^p_n\) &  600,  840 & production capacity for each supplier, respectively \\
               & \(b^f_n\) &  840,  960 & processing capacity for each factory, respectively \\
               & \(b^s_n\) & 6400, 7200 & stock capacity for each factory, respectively \\
               & \(b^s_n\) & 1600, 1800 & stock capacity for each pair of other nodes at the same echelon \\
    \noalign{\smallskip}\hline\noalign{\smallskip}
    Initial values & \(s_{n}\)   &     800 & stock level for all nodes \\
                   & \(p_{in}\)  & 600,840 & material to be available on time steps $1,...,l^{avg}$ on each supplier \\
                   & \(t_{inm}\) & 600,840 & material to arrive on time steps $1,...,l^{avg}$ for each factory  \\
                   & \(t_{inm}\) & 240,240 & idem, but for each wholesaler or retailer \\
    \noalign{\smallskip}\hline    
    \end{tabular}}
\end{center}
\end{table}

We have designed the scenarios to have variety in terms of demand types (seasonal and regular), demand uncertainty, and lead times (stochastic and constant).
In scenarios with seasonal demands, the demand values of each retailer are generated using sinusoidal and perturbation functions.
The sinusoidal function $S$, presented in Equation~\ref{eq:sin_demands}, generates data with seasonal behavior, where $min$ and $max$ represent the minimum and maximum curve values, $z$ is the number of function's peaks, and $t$ is the related time step.
The value of the sinusoidal function is added to a perturbation function $P$ (defined for each scenario) parameterized by an uncertainty level $p$, as shown in Equation \ref{eq:demands}.
In this equation, $min^{sin}=100$ and $max^{sin}=300$ are the minimum and maximum values for the sinusoidal function, and $min=0$ and $max=400$ are the minimum and maximum possible demand values.
If we remove the perturbation term from the equation, we have deterministic seasonal demands, that could be seen as forecast demand values.
In the case of scenarios with non-seasonal (regular) demands, the demand values are generated by $d^{avg}+P(p)$, where $d^{avg}=200$.

\begin{equation} \label{eq:sin_demands}
S(min,max,z,t) = min + \frac{max-min}{2}\Big[1 + \sin{\Big(\frac{2.z.t.\pi}{h}}\Big)\Big]
\end{equation}

\begin{equation} \label{eq:demands}
D = clip\Big(S(min^{sin}, max^{sim}, z, t) + P(p), min, max\Big)
\end{equation}

Regarding scenarios with stochastic lead times, the lead time values of each node are sampled from a Poisson distribution, given by $min(Poisson(l^{avg}-1)+1,l^{max})$, where $l^{avg}=2$ is the average lead time, and $l^{max}=4$ the maximum lead time.
We have used this construction to avoid zero lead times and to keep the lead times in the interval $[1,4]$.
In scenarios with constant lead times, we have used the average value $l^{avg}=2$.

\begin{figure*}
  \includegraphics[width=\textwidth]{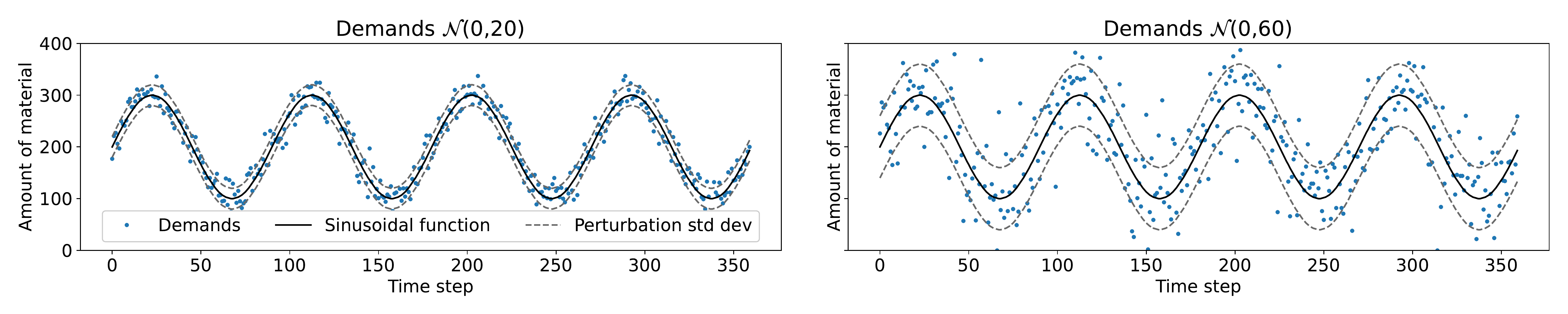}
\caption{Example of demands for scenarios \textit{N20} and \textit{N60}: solid black lines are the sinusoidal function (representing expected values), dashed gray lines represent the standard deviation of the perturbation ($\mathcal{N}(0,20)$ and $\mathcal{N}(0,60)$, respectively), and blue dots show an instance of demands for one retailer.}
\label{fig:demands}       
\end{figure*}

Table \ref{tab:scenarios_param} shows the proposed experimental scenarios, presenting their differences.
Scenarios of set $A$ were designed to verify the behavior of the PPO2 agent considering stochastic lead times and different levels of uncertainty for seasonal demands.
Set $B$ is similar to the first group but considering constant lead times.
In both sets, the uncertainty of the demand values (perturbation) is given by a Gaussian (Normal) distribution, with mean zero and standard deviation $p$.
The values of $p$, from 0 to 60, were chosen to represent different uncertain levels, from no uncertainty to higher levels.
The $p$ value was limited to 60 to ensure that demand values, although uncertain, would remain seasonal.
Figure \ref{fig:demands} shows examples of demands for scenarios \textit{N20} and \textit{N60}.
Black solid lines are the values of the sinusoidal function, without the perturbation.
Gray dashed lines represent the standard deviation of the perturbation ($\mathcal{N}(0,20)$ and $\mathcal{N}(0,60)$, respectively).
Blue dots show examples of demand values for one retailer.
Table \ref{tab:scenarios_param} also shows the sets $C$ and $D$, that contain scenarios with regular demands, considering stochastic and constant lead times, respectively.
Perturbation functions can be given by Gaussian or Uniform distributions.
In the case of scenarios in which demands are generated by Uniform distribution, the demand values are given by $200+\mathcal{U}([-200,200])$, which is the same as saying that they are uniformly sampled from $[0,400]$ interval.
Thus, we have designed the scenarios of sets $C$ and $D$, considering an increasing level of demand uncertainty.
Finally, a stock costs variation is evaluated with the scenario of group $E$.

\begin{table}
\caption{Experimental scenarios and their differences. Scenario \textit{N20stc} is equal to \textit{N20} except that the stock costs are [1,2,1,2,5,6,5,6].}
\label{tab:scenarios_param}
\addtolength{\tabcolsep}{-2pt}
\begin{tabular}{llcccccc}
\hline\noalign{\smallskip}
\textbf{Set} & \textbf{Scenario} & \phantom{abc} & \textbf{Seasonal Demands} & \textbf{Pert. Function} & $p$ & \phantom{abc} & \textbf{Stochastic Lead times}  \\
\noalign{\smallskip}\hline\noalign{\smallskip}
 & \textbf{N0}       & & \checkmark & \textit{none} &     & & \checkmark \\
A & \textbf{N20}     & & \checkmark & $\mathcal{N}$ & 20  & & \checkmark \\
 & \textbf{N40}      & & \checkmark & $\mathcal{N}$ & 40  & & \checkmark \\
 & \textbf{N60}      & & \checkmark & $\mathcal{N}$ & 60  & & \checkmark \\
\noalign{\smallskip}\hline\noalign{\smallskip}
 & \textbf{N0cl}     & & \checkmark & \textit{none} &     & & \\ 
B & \textbf{N20cl}   & & \checkmark & $\mathcal{N}$ & 20  & & \\ 
 & \textbf{N40cl}    & & \checkmark & $\mathcal{N}$ & 40  & & \\ 
 & \textbf{N60cl}    & & \checkmark & $\mathcal{N}$ & 60  & & \\ 
\noalign{\smallskip}\hline\noalign{\smallskip}
  & \textbf{rN0}     & &            & \textit{none} &     & & \checkmark \\
C & \textbf{rN50}    & &            & $\mathcal{N}$ & 50  & & \checkmark \\
  & \textbf{rN100}   & &            & $\mathcal{N}$ & 100 & & \checkmark \\
  & \textbf{rU200}   & &            & $\mathcal{U}$ & [-200,200] & & \checkmark \\
\noalign{\smallskip}\hline\noalign{\smallskip}
  & \textbf{rN0cl}   & &            & \textit{none} &     & & \\ 
D & \textbf{rN50cl}  & &            & $\mathcal{N}$ & 50  & & \\ 
  & \textbf{rN100cl} & &            & $\mathcal{N}$ & 100 & & \\ 
  & \textbf{rU200cl} & &            & $\mathcal{U}$ & [-200,200] & & \\ 
\noalign{\smallskip}\hline\noalign{\smallskip}
E & \textbf{N20stc$^*$}  & & \checkmark & $\mathcal{N}$ & 20  & & \checkmark \\
\noalign{\smallskip}\hline
\end{tabular}
\end{table}

\subsection{Applying PPO2}
\label{subsec:applying_ppo2}

We have chosen the PPO2 algorithm to solve the problem because it achieves high performance in problems with high-dimensional continuous action spaces.
The first step to applying the algorithm is to implement the simulation of the supply chain operation (the environment).
One possible approach would be to implement the environment following exactly the MDP formulation presented in Section \ref{subsec:mdp}. 
But PPO2, and other Deep RL algorithms, usually obtain better results if state and action values are normalized in $[-1,1]$ interval \citep{stable-baselines3}.
In fact, in preliminary experiments, we have tried to apply PPO2 without considering state normalization, and also using automatic state normalization (considering running averages), but the following proposed methodology has obtained better results.

The state values are divided by a maximum limit and then scaled from $[0,1]$ to $[-1,1]$ interval.
Following the notation used for the NLP model, let $b^s_n$ be the node's stock capacity, $b^p_n$ the supplier's capacity, $d^{max}$ the maximum possible demand value, $l^{max}$ the maximum possible lead time value, and $h$ the planning horizon.
The maximum limits used in the state normalization are:
\begin{itemize}
    \item $b^s_n$ for the current stock level of each node.
    \item For each supplier:
    \begin{itemize}
        \item $b^p_n$ for raw material being produced and that will be available in the next time step.
        \item $b^p_n*(l^{max}-1)$ for the sum of raw material being produced and that will be available in the time steps after the next one.
    \end{itemize}
    \item For each other node:
    \begin{itemize}
        \item $b^s_n+b^s_m$ for material in transport arriving at the node in the next time step, delivered by the predecessor nodes $n$ and $m$.
        \item $(b^s_n+b^s_m)*(l^{max}-1)$ for material in transport arriving at the node in the next time step, delivered by the predecessor nodes $n$ and $m$.
    \end{itemize}
    \item $d^{max}$ for customer demands of each retailer.
    \item $h$ for the number of remaining time steps until the end of the episode.
\end{itemize}

Table \ref{tab:state_norm} presents the normalization for some of the values of the example state presented in Figure~\ref{fig:mdp_state}.
In this example we consider $b^s_n=500$, $b^p_n=400$, $l^{max}=4$, and $h=360$.
The first column indicates the type of state value, the second column shows the related time step, and the third column the related node.
Column \textit{Value} shows the actual value in the supply chain simulation, column \textit{Max} the maximum possible value, column \textit{N[0,1]} shows the value divided by the maximum value, and, finally, the last column shows the value scaled for $[-1,1]$ interval.
The last column values are used as the input for the PPO2 algorithm.

\begin{table}
\caption{Partial example of a state normalization for some of the values presented in Figure~\ref{fig:mdp_state}.
The actual values from the supply chain simulation are presented in column \textit{Value} and the values used as input for the PPO2 algorithm are presented in the last column.
}
\label{tab:state_norm}
\begin{tabular}{lllrrrr}
\hline\noalign{\smallskip}
\textbf{Type} & \textbf{Time step} & \textbf{Node} & \textbf{Value} & \textbf{Max.} & \textbf{N[0,1]} & \textbf{S[-1,1]} \\
\noalign{\smallskip}\hline\noalign{\smallskip}
 Stock                & $t$         & Supplier1 & 400 & 500  & 0.800 & 0.600 \\
\noalign{\smallskip}\hline\noalign{\smallskip}
 Raw material         & $t+1$       & Supplier2 & 330 & 400  & 0.825 & 0.650 \\
 being produced       & after $t+1$ & Supplier2 & 105 & 1200 & 0.088 & -0.825 \\
\noalign{\smallskip}\hline\noalign{\smallskip}
 Material             & $t+1$       & Factory1  & 280 & 1000 & 0.280 & -0.440 \\
 in transport         & after $t+1$ & Factory1  & 420 & 3000 & 0.140 & -0.720 \\
\noalign{\smallskip}\hline\noalign{\smallskip}
 Customer demands     & $t+1$       & Retailer1 & 138 & 400  & 0.345 & -0.310 \\
\noalign{\smallskip}\hline\noalign{\smallskip}
 Remaining time steps & $t$         &           & 330 & 360  & 0.916 & 0.833 \\
\noalign{\smallskip}\hline
\end{tabular}
\end{table}

Regarding action values, the output of the PPO2 algorithm is a vector whose values are in the $[-1,1]$ interval.
These values are scaled to $[0,1]$ interval and then changed in the actual values used in the supply chain simulation.
The change from the $[0,1]$ interval to the actual values is done as follows.
Regarding decisions on how much to produce on each supplier, an action value is multiplied by the supplier's capacity ($b^p_n$).
About the decisions related to how much to deliver, let $a_{nm}$ and $a_{no}$ be the action values representing the amount of material to be delivered from a node $n$ to its successor nodes $m$ and $o$.
If the node $n$ is not a factory, the action values are first multiplied by the node's current stock levels $S_{in}$, so that we get $a'_{nm} = a_{nm}S_{in}$ and $a'_{ no} = a_{no}S_{in}$.
In the case of a factory, a treatment is necessary to ensure that the processing capacity of the factory will be respected. 
For this, the stock values are first multiplied by the minimum between the factory's current stock level $S_{in}$ and its processing capacity $b^f_n$, so that we get $a'_{nm} = a_{nm}min(S_{in},b^f_n )$ and $a'_{no} = a_{no}min(S_{in},b^f_n)$. 
The calculated values, $a'_{nm}$ and $a'_{no}$, are used to define the minimum and maximum cuts in material in stock at the node $n$, given by $c^{min}=min(a'_{nm},a'_{no})$ and $c^{max}=max(a'_{nm},a'_{no})$, respectively.
The value $c^{min}$ indicates the amount of material to be delivered to node $k$, such that $c^{min}=a'_{nk}$.
For the other node, the amount of material is given by $c^{max}-c^{min}$.
The remaining material $S_{in} - c^{max}$ is kept in the stock of the node $n$.
With this approach, all possible output values generated by the PPO2 represent feasible action values in the supply chain simulation.

\begin{figure*}
  \includegraphics[width=0.5\textwidth]{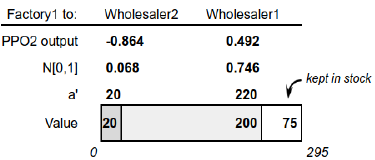}
\caption{Example of the action values regarding how much material to be delivered from Factory1 to the wholesalers.
The output values from PPO2 are scaled to $[0,1]$ interval, then multiplied by Factory1's stock level.
The resulting values are then sorted and viewed as cuts in Factory1's stock.}
\label{fig:action_deliver}
\end{figure*}

To illustrate how the action values are handled, let's use some of the decisions of the action example presented in Figure~\ref{fig:mdp_action}, considering they are taken after observing the state example presented in Figure~\ref{fig:mdp_state}.
In the example, the decision regarding how much to produce in Supplier1 is $210$.
This value would come from an output value $a=0.050$ from PPO2, which would be scaled to the $[0,1]$ interval as $a'=\frac{a+1}{2}=0.525$.
Finally, the decision value would be calculated as $a' b^p_n = 0.525 * 400 = 210$.
Figure~\ref{fig:action_deliver} shows how to handle the delivery decisions from node Factory1.
The output values from PPO2 would be $0.492$ and $-0.864$ for Wholesaler1 and Wholesaler2, respectively.
These values would be scaled for $[0,1]$ interval as $0.746$ and $0.068$, and then multiplied by the stock level ($b^s_n=15+280=295$) of the Factory1, achieving $220$ and $20$, respectively.
The minimum value $c^{min}=20$ indicates that the amount of material to be delivered to Wholesaler2 is $20$.
For the Wholesaler1 the amount is $c^{max}-c^{min}=220-20=200$ units.
The remaining material, $b^s_n - c^{max} = 295-220=75$, would be kept in stock.

We have used preliminary exploratory experiments to take some other decisions regarding the way we have used PPO2.
First, we have decided to use automatic reward normalization (considering running averages).
As mentioned about state and action spaces, it is also advisable to normalize the rewards \citep{stable-baselines3}, and our preliminary experiments confirmed that this approach obtains better results with PPO2.
We have also experimented to give feedback to the agent only at the end of the episode, i.e., consider costs equal to zero at every time step except the last one, in which the reward is the total accumulated cost.
The idea is that we intend to minimize the total operation costs, and it does not matter how the costs are allocated over the planning horizon.
But the results were not better with this approach, so we keep the rewards as the costs incurred in each time step.

The proposed methodology could be adapted to work with discrete state and action spaces or unlimited capacities.
In the case of discrete values, one possible approach would be to keep the states and actions as continuous values from the agent's point of view but rounding the action values before using them in the supply chain simulation.
Another possible approach would be to use discrete state and action spaces, as PPO2 can work in this setting as well, but we believe that the first approach would obtain better results.
In the case of unlimited capacities, one could use simulation to find upper bounds for the state and action values that would make sense with the customer demands' range.
Then the upper bounds could be used to normalize the values in a similar way we have done in our methodology.

\subsection{The Baseline: LP Agent}
\label{subsec:lpagent}

We have decided to use an LP-based agent as a baseline in our experiments.
Although related works usually use heuristics as baselines, like $(r,Q)$ or base stock, we believe that they are better suited for order-based approaches, or when the chain is linear.
In our case, with a non-serial supply chain, it is not so easy to adapt such heuristics, since it would be a trick to define how to combine the decisions of two nodes on each echelon.
In fact, the only related work that deals with non-serial supply chains \citep{Perez2021} also uses an LP-based baseline.
Furthermore, as we handle the problem with a production planning approach, we believe that an LP-based baseline using forecast demands and average lead times is a more practical approach.

As mentioned in Section~\ref{subsec:nlp}, the presented NLP model becomes an LP model if we consider forecast demand values and average lead times.
In the experiments, seasonal demands are generated from sinusoidal and perturbation functions, as presented in Section~\ref{subsec:scenarios}.
So we can consider the sinusoidal function without the perturbation as a forecast value for the demands.
In scenarios with regular demands, the forecast value can be defined as the average demand.
Thus, we can solve the LP model considering such forecast demand values and average lead times with an LP solver.
The LP model's solution is then used to encode the LP agent employed as a baseline in the experiments.

The LP agent is built from the decision variables $P_{ijn}$ and $T_{ijnm}$ of the model (production at the suppliers and transport of material, respectively).
The other decision variables (stock, material processed, excess of material, and missing products) do not need to be encoded in the agent, since they will be handled by the simulation of the supply chain.
The values of the used decision variables are normalized and scaled to $[-1,1]$ interval.
Thus, the LP agent can interact with the environment in the same way as the PPO2 agent does.

\subsection{Lower Bounds: Perfect Information LP}
\label{subsec:lower_bounds}

The LP model can also be used to find lower bounds for each experimental scenario, by solving the problem after the fact with perfect information.
If we solve the model using the true demand and lead times values, as if they were known in advance, we obtain the lower bounds for the total operating costs.
The model solution is optimal considering the realization of the demand and lead time values, but it has lower total costs than the optimal value of the original problem with stochastic demands and lead times.
Nevertheless, it can be viewed as an oracle for benchmark purposes.

\subsection{Experimental Methodology}
\label{subsec:exp_meth}

\begin{figure*}
  \includegraphics[width=0.6\textwidth]{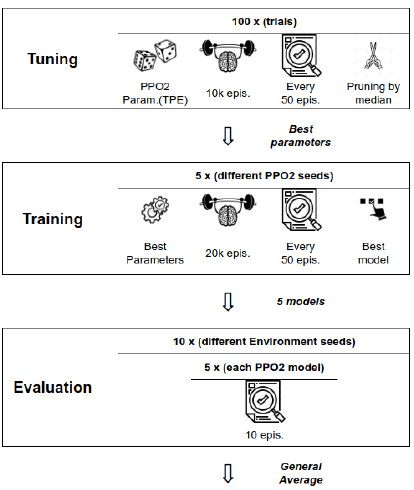}
\caption{The experimental methodology consists of three phases: tuning of the PPO2 hyperparameters, training using the best parameter values found, and evaluation of the results using the best PPO2 models.}
\label{fig:methodology}       
\end{figure*}

The experimental methodology consists of three parts: hyperparameter tuning, training, and evaluation and it is illustrated in Figure~\ref{fig:methodology}.

The first part, hyperparameters tuning, is essential to obtain better results with Deep RL algorithms and is present in the top part of Figure~\ref{fig:methodology}.
The proposed methodology uses 100 different combinations (trials) of hyperparameter values.
At the first 20 attempts, the values are randomly chosen from predefined intervals.
The remaining 80 attempts use the TPE algorithm \citep{bergstra2011algorithms} to choose the parameter values.
For each combination of hyperparameter values, i.e., each attempt, the agent is trained for 3.6 million time steps (equivalent to 10 thousand episodes), with evaluations of the model on every 50 episodes (18 thousand time steps).
Each evaluation step consists of 5 episodes, and the average of the accumulated rewards is used to define the quality of the model.
The best model found in all evaluations is considered as the result of the attempt.
A pruning by median mechanism is used in the last 80 attempts.
Unpromising attempts are early-stopped using the median stopping rule, that is, an attempt is pruned if the intermediate result of the trial is worse than the median of the previous trials.
Finally, the values of the parameters used in the attempt with the best results are chosen for the experiments.
The hyperparameter tuning is done considering one scenario, and the resulting parameter values are used for all experimented scenarios.

After selecting the PPO2 hyperparameter values, the second part of the methodology refers to training the PPO2 agent (middle part of Figure~\ref{fig:methodology}).
It is important to repeat the training considering different random seeds to ensure the robustness of the results achieved by RL algorithms \citep{henderson2019deep}.
Therefore we train the PPO2 agent five times, with different predefined random seeds, for each experimental scenario.
Each training consists of running the algorithm for 7.2 million time steps\footnote{In preliminary experiments, we have tried training runs with 3.6 million time steps but, for several scenarios, the model was still being improved until the end of the training. So, we have decided to run for 7.2 million time steps, and we have verified that the model stopped being improved before the end of the training.}
(or 20 thousand episodes), evaluating the model on every 50 episodes (or 18 thousand time steps), considering 10 episodes on each evaluation step.
The best model found on each training, i.e., for each random seed, is used to evaluate the results.

Finally, the evaluation of the results is the third part of the experimental methodology and is presented in the bottom part of Figure~\ref{fig:methodology}. 
The evaluation consists of simulating 100 episodes of the environment for each PPO2 model found in the training process. 
The 100 episodes are generated using 10 different predefined random seeds for the environment.
With this approach, a total of 500 episodes of evaluation is planned to be executed for each scenario, and the resulting metric is the average and standard deviation of accumulated rewards of all these episodes.

The LP agent, presented in Section \ref{subsec:lpagent}, is used as a baseline to be compared to PPO2.
The agent is built for each scenario from the solution of the LP model.
The model is solved considering average lead times and forecast (average) demands.
As presented, forecast demands mean generating the demands without the perturbation term.
To evaluate the LP agent, we have used the same 100 episodes of the environment used with PPO2, i.e., the same sequence of demands and lead times for each episode.
Therefore, we compare the results obtained by PPO2 and the baseline, considering the same conditions.

\section{Experiments and Results}
\label{sec:experiments}

The experimentation was conducted using Python 3.6.10 on a computer with a 2.9 GHz x 6 processor, 32 GB of RAM, a 6 GB GPU, and Ubuntu Linux 20.04.
The supply chain simulation was implemented in Python following the OpenAI Gym standard \citep{OpenAIGym} and the LP model was solved using CPLEX 12.10 via Python interface.
The PPO2 version of the Stable Baselines 3 (SB3) library \citep{stable-baselines3} was used in the experiments, and the RL Baselines3 Zoo library\footnote{RL Baselines3 Zoo library is a wrapper to use Optuna library \citep{akiba2019optuna} with SB3.} \citep{rl-zoo3} was used for the hyperparameter tuning. 
We started the experiments with Stable Baselines 2 \citep{stable-baselines}, but SB3 has achieved better results in preliminary experiments. 
Although we have verified that default hyperparameter values were the main reason for the difference between the results of the two versions of the library, we have chosen SB3 as its authors state that PPO2 implementation in this version is closer to the original one.

The remainder of this section is organized as follows.
In Section \ref{subsec:tuning}, we present the hyperparameter tuning phase.
Section \ref{subsec:eval_results} presents the main results for all experimented scenarios.
Section \ref{subsec:stocks_seasonality} shows that, in the scenarios with seasonal demands, the PPO2 agent can build stocks in advance, with seasonality.
Next, in Section \ref{subsec:cost_types}, the analysis of the results is detailed to verify the performance of the PPO2 agent regarding each type of cost.
The learning curves of the PPO2 are discussed in Section \ref{subsec:learning_curve}.
Finally, in Section~\ref{subsec:summary}, we present a summary of the results.

\subsection{Hyperparameter Tuning}
\label{subsec:tuning}

Following the proposed methodology (Section \ref{subsec:exp_meth}), we start with hyperparameter tuning using scenario \textit{N20}.
Table \ref{tab:param_tuning} shows the best values found for the PPO2 hyperparameters.
It is presented for each hyperparameter: the type of sampling (categorical, uniform, or log uniform), the predefined possible values (or interval), the best value found at the end of the tuning process, and the description of the hyperparameter.
Regarding the parameters with fixed values, we have experimented with four actors, and the \texttt{ent\_coef} parameter was fixed to zero since it is intended to be used with discrete action spaces \citep{stable-baselines3}.
The predefined possible values were chosen from preliminary tuning experiments.
We have started with default values of the RL Baselines3 Zoo and then and we reduced the options of some parameters to the values that achieved the best results.
The code of RL Baselines 3 Zoo library was modified to use the chosen predefined possible values and to fix the values of the first attempt\footnote{The first of the 100 trials was fixed to use the SB3 default hyperparameter values for PPO2, as the library documentation states that they are optimized for continuous problems \citep{stable-baselines3}.}.
Regarding optimizer, we have used the Adam method for gradient optimization \citep{kingma2017adam}.
The training, the second step of the methodology, of all proposed scenarios were done using the best hyperparameter values found in the tuning process.

\begin{table}
    \caption{Best PPO2 parameter values found in hyperparameter tuning; columns S indicates the sampling options, which are C: categorical, U: sampled from the interval in the linear domain, L: sampled from the interval in the log domain, and -: fixed}
    \label{tab:param_tuning}
    \begin{center}
    \resizebox{\textwidth}{!}{%
        \begin{tabular}{lllll}
        \hline\noalign{\smallskip}
        \textbf{Parameter} & \textbf{S.} & \textbf{Possible values} & \textbf{Best value} & \textbf{Description}  \\
        \noalign{\smallskip}\hline\noalign{\smallskip}
        \texttt{n\_steps}        & C & $2^{5}, 2^{6}, ..., 2^{11}$ & 1,024 & $\tau$ parameter in Algorithm \ref{alg:PPO} \\
        \texttt{n\_epochs}       & C & [3, 5, 10, 20] & 20 & $K$ parameter in Algorithm \ref{alg:PPO} \\
        \texttt{batch\_size}     & C & [64, 128, 256, 512] & 64 & Mini-batch size in Algorithm \ref{alg:PPO} \\
        \texttt{vf\_coef}        & U & [0, 1] & 0.88331 & $c_1$ in Equation \ref{eq:PPO_obj_function} \\
        \texttt{clip\_range}     & C & [0.1, 0.2, 0.3] & 0.2 & $\epsilon$ in Equation \ref{eq:PPO_Lclip} \\
        \texttt{gae\_lambda}     & C & [0.9, 0.92, 0.95, 0.98, 1.0] & 0.95 & $\lambda$ used to calculate GAE \\
        \texttt{gamma}           & C & [0.95, 0.98, 0.99,  0.995, 0.999, 0.9999] & 0.999 & $\gamma$ used to calculate GAE \\
        \texttt{net\_arch}       & C & [(64,64), (128,128), (256,256)] & (64,64) & Units of ANN's hidden layers \\
        \texttt{lr\_schedule}    & C & [constant, linear] & constant  & Learning rate schedule \\
        \texttt{learning\_rate}  & L  & [0.00001, 0.01] & 0.0001 & Gradient method's step size \\
        \texttt{activation\_fn}  & C & [ReLU, TanH] & TanH & ANN's activation function \\
        \texttt{max\_grad\_norm} & C & [0.3, 0.5, 0.6, 0.7, 0.8, 0.9, 1, 2, 5] & 0.5 & To clip normalized gradients \\
        \noalign{\smallskip}\hline\noalign{\smallskip}
        \texttt{n\_actors}  & - & 4 & 4 & $N$ in Algorithm \ref{alg:PPO} \\
        \texttt{ent\_coef} & - & 0 & 0 & $c_2$ in Equation \ref{eq:PPO_obj_function} \\
        \noalign{\smallskip}\hline\noalign{\smallskip}
        \end{tabular}}
    \end{center}
\end{table}

\subsection{Main Results}
\label{subsec:eval_results}

Table \ref{tab:scenarios_results} summarizes the results of the experiments by comparing PPO2 and LP agents for all scenarios (detailed results are available in Online Resource 1).
The first column shows the set of scenarios, the second column indicates whether demands are seasonal or regular, and the third column whether lead times are constant or stochastic.
The fourth column shows the scenario's name, the fifth and sixth columns present the lower bounds for the total operating costs (i.e., optimal solution if demands and lead times were known in advance).
The following four columns present the average and the standard deviation of the total operating costs for the LP and PPO2 agents.
The last two columns show the gain of PPO2 over the  LP agent (difference and percentage, respectively).

\begin{table}
\caption{Results for all considered scenarios: each set of scenarios indicates whether demands are seasonal or regular and whether lead times are constant or stochastic. 
The table presents, for each scenario, the average and standard deviation of the total operating costs regarding the lower bounds, LP-agent (the baseline), and PPO2 agent.
The last two columns present the gain of PPO2 over LP (difference and percentage, respectively).
The numbers in the scenarios' names indicate the perturbation of the demand.}
\label{tab:scenarios_results}
\addtolength{\tabcolsep}{-3pt}
\begin{tabular}{lcclrrrrrrrr}
\hline\noalign{\smallskip}
\textbf{Set} & \textbf{Seas.} & \textbf{Stoch.} & \textbf{Scenario} & \multicolumn{2}{c}{\textbf{Lower Bound}} &  \multicolumn{2}{c}{\textbf{LP agent}} & \multicolumn{2}{c}{\textbf{PPO2 agent}} & \multicolumn{2}{c}{\textbf{Gain}} \\
 & \textbf{Dem.} & \textbf{Lead T.} & & \textit{Avg} & $\sigma$ & \textit{Avg} & $\sigma$ & \textit{Avg} & $\sigma$ & value & \% \\
\noalign{\smallskip}\hline\noalign{\smallskip}
 & \checkmark & \checkmark & \textbf{N0}   & 8,004 k & 27 k & 10,298 k & 195 k & 9,147 k & 125 k & 1,151 k & 11.2 \% \\ 
\textbf{A} & \checkmark & \checkmark & \textbf{N20}   & 8,005 k & 49 k & 10,316 k & 207 k & 9,252 k & 157 k &  1,065 k & 10.3 \% \\
  & \checkmark & \checkmark & \textbf{N40} & 8,008 k & 88 k & 10,393 k & 237 k & 9,492 k & 196 k & 901 k & 8.7 \% \\
  & \checkmark & \checkmark & \textbf{N60} & 8,010 k & 128 k & 10,503 k & 276 k &  9,737 k & 206 k & 766 k &   7.3 \% \\
  \noalign{\smallskip}\hline\noalign{\smallskip}
 & \checkmark & & \textbf{N0cl}   & 7,941 k & 0 &  7,941 k & 0 & 8,017 k & 7 k & -76 k & -1.0 \% \\ 
\textbf{B} & \checkmark & & \textbf{N20cl}  & 7,944 k & 42 k &  8,226 k & 61 k & 8,231 k & 80 k & -6 k & -0,1 \% \\
  & \checkmark & & \textbf{N40cl} & 7,951 k & 84 k &  8,501 k & 118k & 8,478 k & 162 k & 22 k & 0.3 \% \\
  & \checkmark & & \textbf{N60cl} & 7,958 k & 124 k &  8,740 k &  171 k & 8,720 k & 201 k & 20 k &   0.2 \% \\
\noalign{\smallskip}\hline\noalign{\smallskip}
  & & \checkmark & \textbf{rN0} & 7,806 k & 8 k & 9,405 k & 142 k & 8,565 k & 49 k & 840 k & 8.9 \% \\
\textbf{C} &  & \checkmark & \textbf{rN50}  & 7,804 k & 91 k &  9,557 k &  257 k & 8,811 k & 124 k & 746 k &  7.8 \% \\
  & & \checkmark & \textbf{rN100} & 7,808 k & 174 k &  9,941 k &  388 k & 9,104 k & 235 k & 837 k & 8.4 \% \\
  & & \checkmark & \textbf{rU200} & 7,817 k & 262 k &  10,143 k & 486 k & 9,219 k &  303 k & 924 k &   9.1 \% \\
\noalign{\smallskip}\hline\noalign{\smallskip}
  & & & \textbf{rN0cl} & 7,652 k & 0 & 7,652 k & 0 k & 7,778 k & 3 k & -126 k & -1.6 \% \\
\textbf{D} &  &  & \textbf{rN50cl}  & 7,647 k & 89 k &  8,283 k & 130 k & 8,098 k &  93 k & 185 k &   2.2 \% \\
  & & & \textbf{rN100cl} & 7,666 k & 173 k &  8,747 k & 240 k & 8,402 k & 180 k & 345 k &   3.9 \% \\
  & & & \textbf{rU200cl} & 7,714 k & 256 k &  8,985 k & 308 k & 8,565 k & 198 k & 420 k & 4.7 \%  \\
\noalign{\smallskip}\hline\noalign{\smallskip}
\textbf{E} & \checkmark & \checkmark & \textbf{N20stc}   & 8,706 k & 101 k & 12,685 k & 249 k & 11,673 k & 243 k & 1,012 k & 8.0 \% \\
\noalign{\smallskip}\hline\noalign{\smallskip}
\end{tabular}
\end{table}

Scenarios of set $A$ have stochastic lead times and seasonal demands.
As can be seen in Table \ref{tab:scenarios_results}, PPO2 agent gain above LP agent is between 7.3\% and 11.2\%. 
Regarding scenarios of set $B$, they have constant lead times and seasonal demands.
In this case, the PPO2 agent is pretty close to the LP agent, with a difference between -1.0\% and 0.3\%.
It is interesting to notice that in scenario \textit{N0cl}, which has no uncertainty, the LP agent result is, in fact, an optimal value.
Therefore in this scenario PPO2 agent achieves a 1.0\% optimality gap.
PPO2 agent performs better than the baseline in scenarios of set $C$, with regular demands and stochastic lead times, by 7.8\% to 9.1\%.
It is also better in scenarios of set $D$, which have regular demands but with constant lead times, by 2.2\% to 4.7\% (except in scenario \textit{rN0cl}, which has no uncertainty, and therefore LP agent achieves an optimal value).
Finally, PPO2 is also better in the scenario of set $E$, which has different stock costs.
This scenario is similar to \textit{N20}, so demands are seasonal and lead times stochastic, and PPO2 gain is 8.0\%.

\begin{figure*}
  \includegraphics[width=\textwidth]{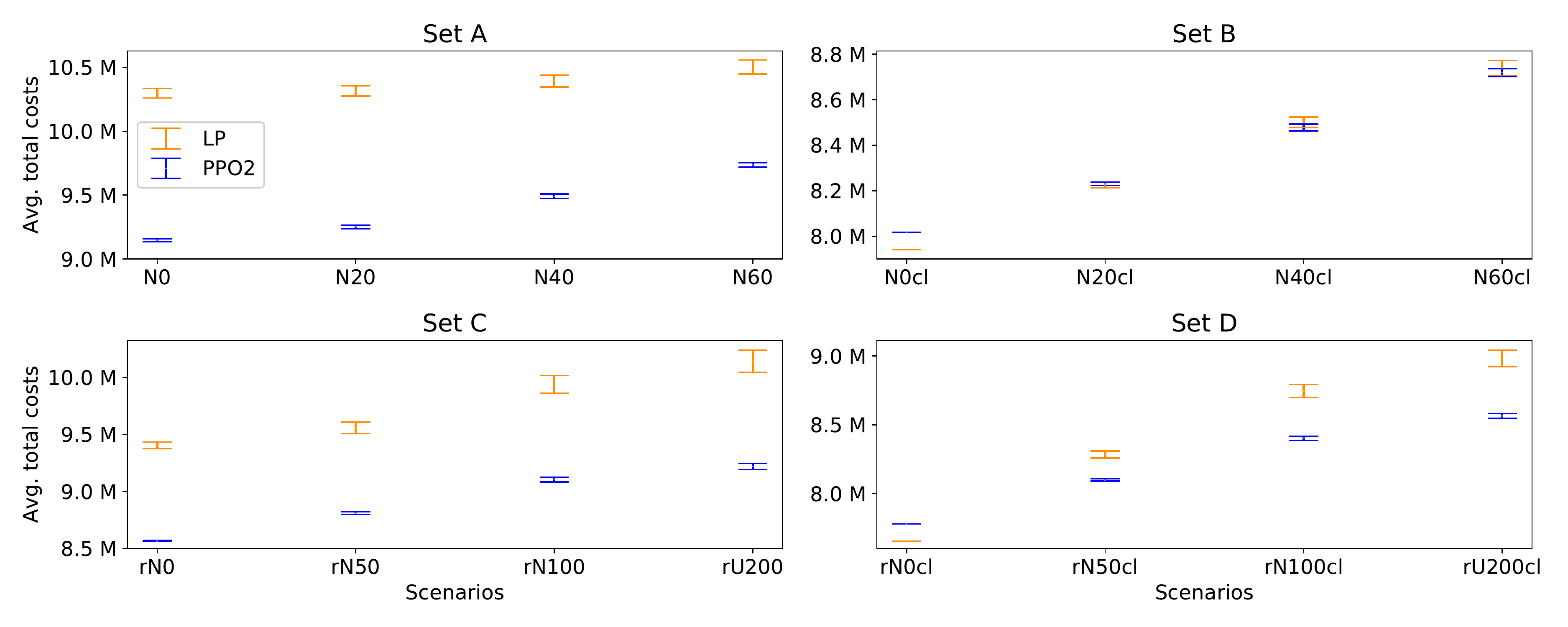}
\caption{95\% confidence intervals for average results of PPO2 and LP agents, obtained via bootstrapped sampling (with 10 k iterations and pivotal method).
The intervals overlap only in scenarios of set B.
In the scenarios of the other sets, the difference between the results achieved by PPO2 and LP are significant.
}
\label{fig:results_CI}       
\end{figure*}

Figure \ref{fig:results_CI} shows 95\% confidence intervals of average results obtained by PPO2 and LP agents.
We have used bootstrapped sampling \citep{efron1994introduction}, with 10 k iterations and the pivotal method, to generate statistically relevant confidence intervals, as suggested by \cite{henderson2019deep}.
We find that PPO2 has small confidence bounds from the bootstrap, showing that the mean value is representative of the performance of the algorithm.
The PPO2 confidence intervals are smaller than LP agent intervals, but this is expected since we have more data for PPO2 (as we evaluate the same 100 episodes for both agents, but for PPO2 is done for each of the 5 resulting models).
Regarding the distance between the PPO2 and LP intervals, we can see that only in scenarios of set B (constant lead times and seasonal demand) the confidence intervals of both agents overlap.
In the scenarios of the other sets, the difference is significant, especially in scenarios of sets A and C.
Considering set D (except scenario \textit{rN0cl} that has no uncertainty), although the difference between the agents is quite small (2.2 to 4.7\%), the confidence intervals do not overlap, so we can say that PPO2 is statistically better than LP.

If we look at the results focusing on constant vs. stochastic lead times, we can see that PPO2 is a good tool to use with stochastic lead times.
The technique is better than baseline in all 9 scenarios with uncertain lead times (between 7.3\% and 11.2\%), regardless of whether demands are seasonal or not. 
PPO2 is also better considering constant lead times if demands are regular with bigger uncertainty (2.2\% to 4.7\%).
Considering constant lead times and seasonal demands, PPO2 achieves pretty the same level of performance as baseline (between -1.0\% and 0.3\%).

Now, focusing on demands, we can see that the costs (and variance) of the PPO2 agent grow with the level of uncertainty, as it would be expected.
In scenarios with regular demands or constant lead times, the difference between PPO2 and LP agents, in general, also grows with the uncertainty of the demands.
In scenarios with seasonal demands and stochastic lead times, the difference between the two agents decreases with the uncertainty of the demands.

\subsection{Stocks with Seasonality}
\label{subsec:stocks_seasonality}

\begin{figure*}
  \includegraphics[width=\textwidth]{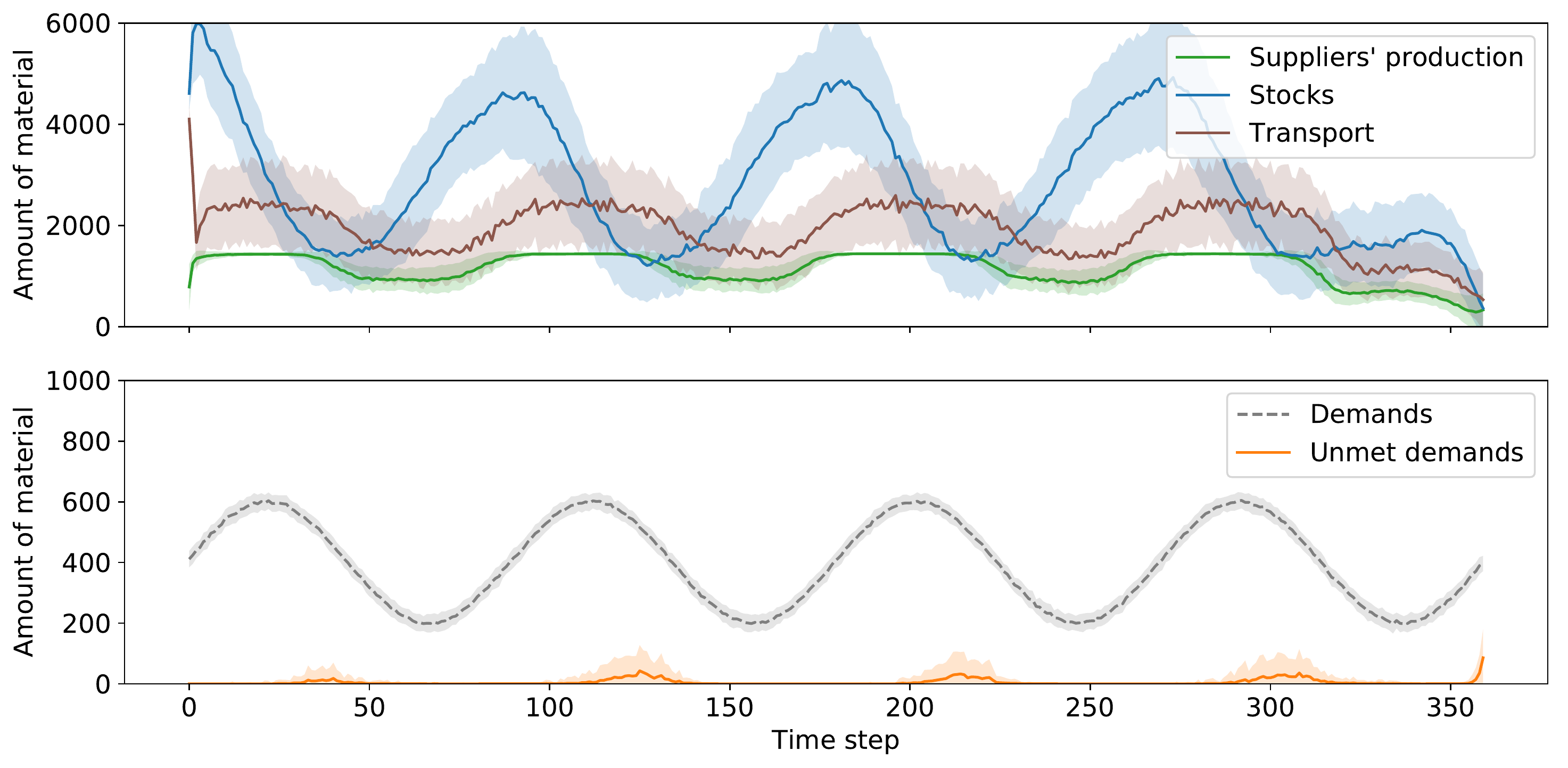}
\caption{The average amount of material by type and by time step considering all evaluated episodes for scenario \textbf{N20}. 
Shaded areas denote $\pm 1$ standard deviation, and values refer to the sum for all nodes of the chain.
Stocks follow the sinusoidal pattern of the demands.}
\label{fig:n20_by_step}       
\end{figure*}

\begin{figure*}
  \includegraphics[width=\textwidth]{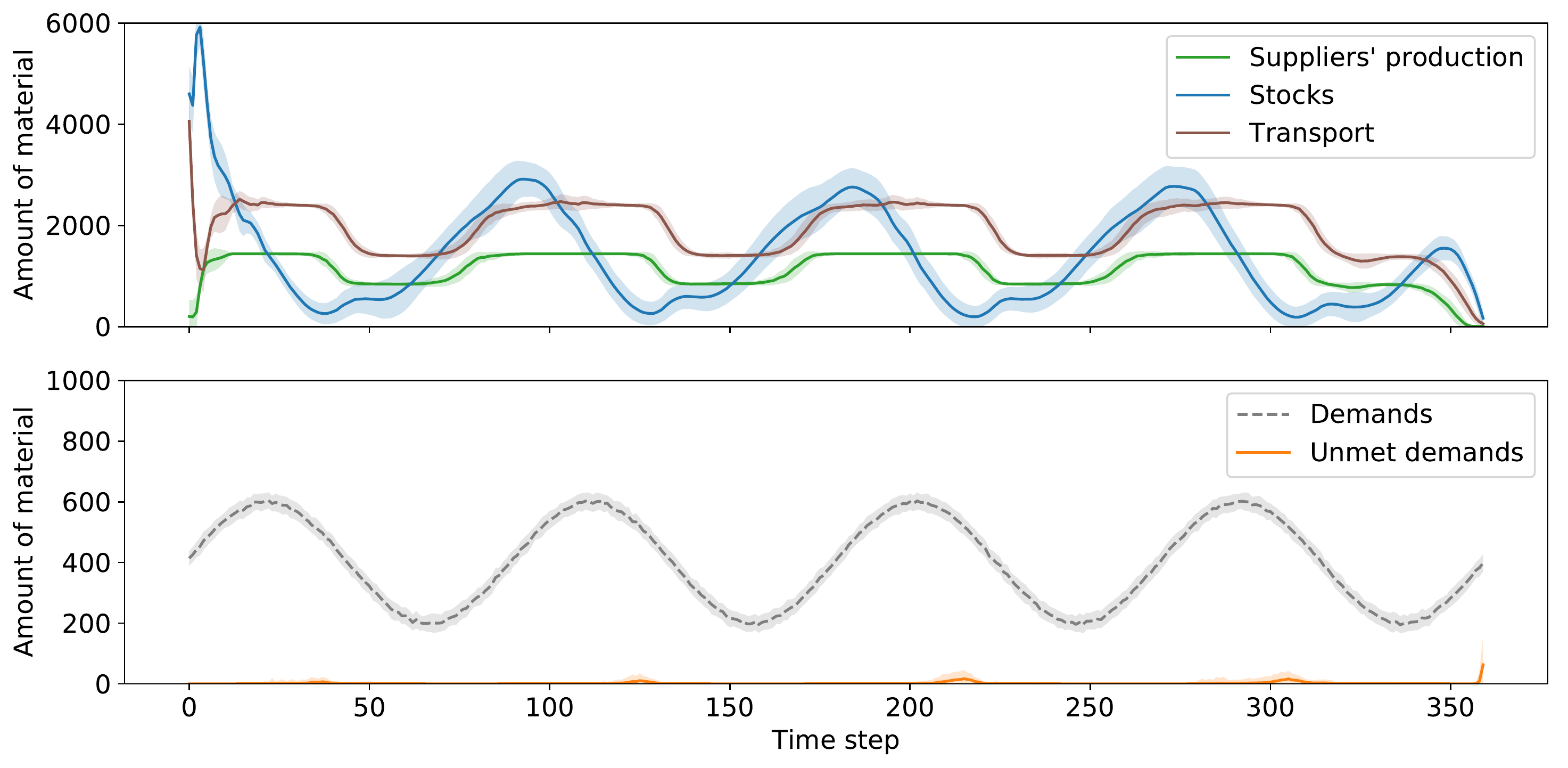}
\caption{The average amount of material by type and by time step considering all evaluated episodes for scenario \textit{N20cl}. Shaded areas denote $\pm 1$ standard deviation. Behaviour is similar to that observed for scenario \textit{N20} in Figure \ref{fig:n20_by_step}, but with a lower level of stocks and lower variance.}
\label{fig:n20cl_by_step}       
\end{figure*}

In this research, we are using the PPO2 algorithm to solve the addressed problem considering seasonal demands. 
Thus it is interesting to evaluate if the agent can build stocks with seasonality.
Figure \ref{fig:n20_by_step} shows different types of costs over the planning horizon regarding scenario \textit{N20}.
The values are the average of all evaluated episodes and they refer to the sum for all nodes of the chain.
The top graph shows production at suppliers, stock and transport costs and the bottom graph shows unmet demands cost.
The sum of the demands for both retailers is also included in the bottom graph for reference.
We can see that stocks are built following the demand sinusoidal pattern with a small shift, showing that the PPO2 agent starts to build stocks before demands start rising.
Graphs show also that unmet demands occur when stocks achieve their lowest levels.
It can be noted that production at suppliers and transport of material also follow the demands sinusoidal pattern.
Another evaluation is that PPO2 was able to decrease production at the end of the episode, as production at suppliers, stock, and transport costs decay at the final time steps.
Figure \ref{fig:n20cl_by_step} shows the same type of data for scenario \textit{N20cl} and the behavior is similar to that observed for scenario \textit{N20}.
Comparing both scenarios, we can see that, to manage the uncertainty of the lead times, the level of stocks in scenario \textit{N20} is higher than in scenario \textit{N20cl}.
Another difference is that the variance of the curves is bigger in the scenario with stochastic lead times.
Finally, PPO2 can better attend to customer demands in scenario \textit{N20cl}, with constant lead times.
Other scenarios with seasonal demands also have similar behaviors, but they are not reported here for the sake of space.

\subsection{Types of Cost}
\label{subsec:cost_types}

\begin{figure*}
  \includegraphics[width=0.9\textwidth]{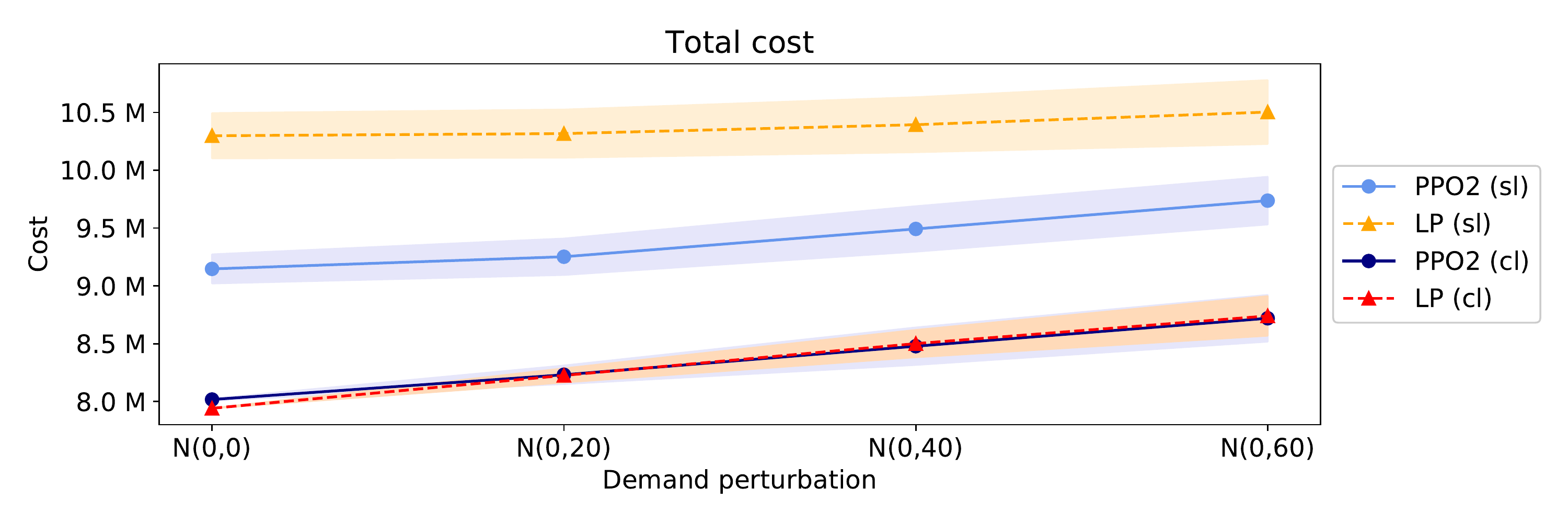}
\caption{Results for scenarios of sets $A$ and $B$ (seasonal demands); 
\textit{sl} means stochastic lead times and \textit{cl} constant lead times. 
The horizontal axis refers to the level of demand perturbation, and the vertical axis refers to the total operating costs.
Shaded areas denote $\pm 1$ standard deviation.
LP agent is represented by dashed lines and triangular markers while PPO2 by solid lines and circular markers.
PPO2 is pretty close to LP agent considering constant lead times and is better with stochastic lead times.}
\label{fig:seasonal_totalcosts}       
\end{figure*}

\begin{figure*}
  \includegraphics[width=\textwidth]{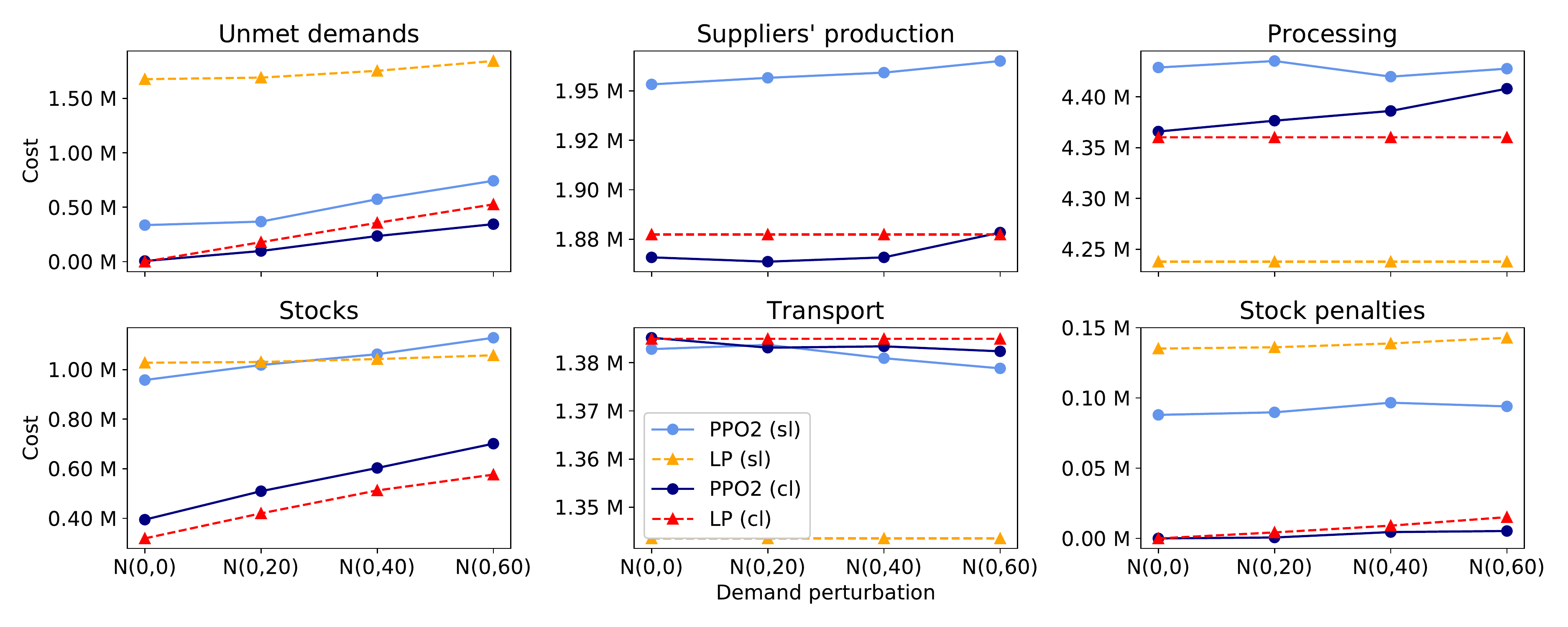}
\caption{Costs by type for scenarios of sets $A$ and $B$ (seasonal demands);
\textit{sl} means stochastic lead times and \textit{cl} constant lead times.
The horizontal axis refers to the level of demand perturbation and, the vertical axis refers to the total operating costs by type.
LP agent is represented by dashed lines and triangular markers while PPO2 by solid lines and circular markers.}
\label{fig:seasonal_costsbytype}       
\end{figure*}

Another valuable analysis is to investigate which types of costs are responsible for the difference between the agents.
Let's evaluate this in the scenarios with seasonal demands of sets A and B, whose final results are presented in Figure \ref{fig:seasonal_totalcosts}.
As mentioned, PPO2 is better than LP agent in scenarios with stochastic lead times and has roughly the same level of performance regarding constant lead times.
Figure \ref{fig:seasonal_costsbytype} shows the composition of the final costs for each scenario, considering each type of cost.
The values are the average of all evaluated episodes.
It is important to notice that the vertical axis of each graph has a different range.
First, considering stochastic lead times, we can see that the main reason why PPO2 has lower costs is that it is more efficient in meeting customer demands.
PPO2 can better meet demands due to two reasons: by producing more material (leading to bigger operations costs related to production at suppliers, process and transport of material); and by having less material discarded by the excess of materials in stocks (or, saying in other words, by respecting better the stock capacities).
We can also see that, in the case of the PPO2 agent, the level of stock grows with demand perturbation.

Now, let's analyze the scenarios with constant lead times.
As demand uncertainty grows, both agents lose more customer demands, but PPO2 becomes more efficient than LP agent (PPO2 is worst only in the scenario with no demand perturbation, in which LP agent has an optimal value).
However, to better meet demands, the PPO2 agent needs to build bigger stocks, while other operations costs are pretty similar.
In real-world scenarios, with uncertain seasonal demands and constant lead times, PPO2 would be a more viable option if the level of uncertainty for the demands is high, or if it is difficult to model the problem (or even to get accurate values for its parameters).

\begin{figure*}
  \includegraphics[width=0.9\textwidth]{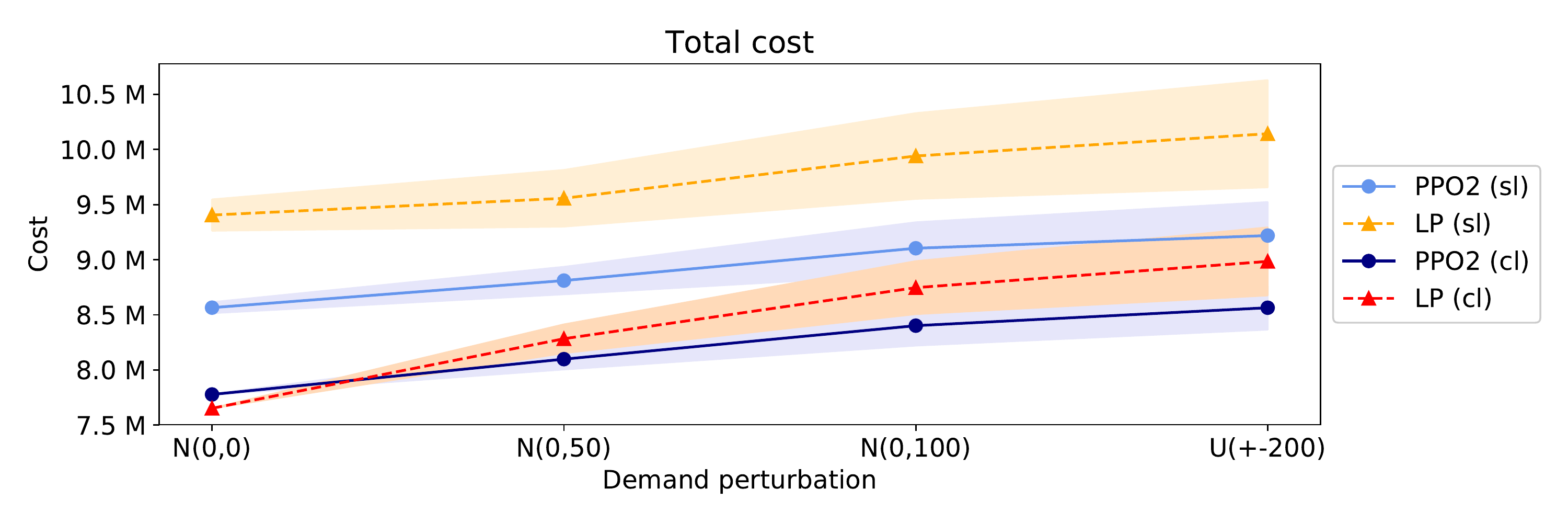}
\caption{Results for scenarios of sets $C$ and $D$ (regular demands);
\textit{sl} means stochastic lead times and \textit{cl} constant lead times.
The horizontal axis is the level of demand perturbation and, the vertical axis is the total operating costs.
Shaded areas denote $\pm 1$ standard deviation.
LP agent is represented by dashed lines and triangular markers while PPO2 by solid lines and circular markers.
PPO2 is better than LP in all scenarios (except \textbf{N0cl} where there is no uncertainty, and the LP agent achieves an optimal value), and the difference is higher with stochastic lead times.}
\label{fig:regular_totalcosts}       
\end{figure*}

\begin{figure*}
  \includegraphics[width=\textwidth]{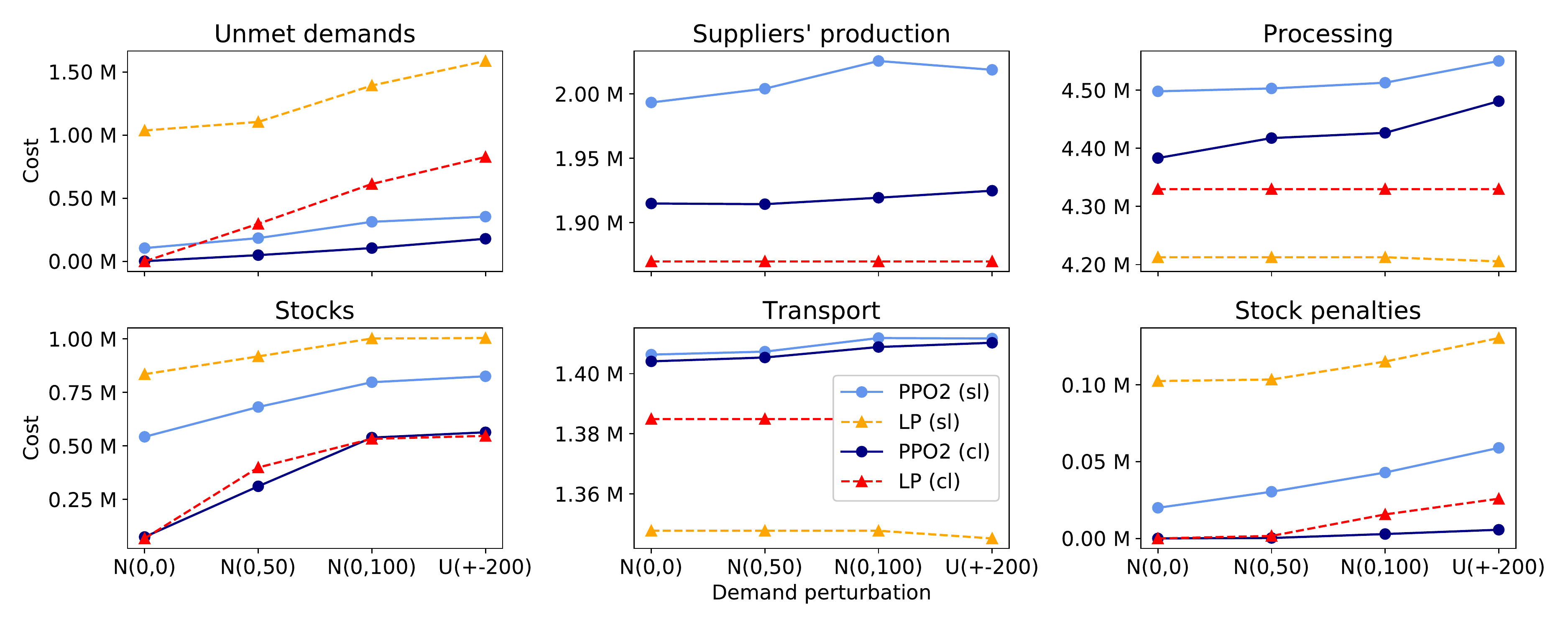}
\caption{Costs by type for scenarios of sets $C$ and $D$ (regular demands); 
\textit{sl} means stochastic lead times and \textit{cl} constant lead times.
The horizontal axis is the level of demand perturbation and, the vertical axis is the total operating costs by type.
LP agent is represented by dashed lines and triangular markers while PPO2 by solid lines and circular markers.}
\label{fig:regular_costsbytype}       
\end{figure*}

The same type of analysis, regarding types of cost, was also done for scenarios of sets $C$ and $D$, considering regular demands.
Figure \ref{fig:regular_totalcosts} shows the final results, i.e., the total operating costs for each scenario, comparing PPO2 and LP agents.
PPO2 is better than baseline in almost all scenarios, and the greater the uncertainty bigger the difference.
LP agent is better only in scenario \textit{N0cl}, in which there is no uncertainty and, therefore, LP agent achieves an optimal value.
The comparison regarding the type of cost is presented in Figure \ref{fig:regular_costsbytype}.
PPO2 agent better attends to customer demands in all scenarios and has lower levels of stock penalties (except in scenario \textit{N0cl}).
PPO2 agent achieves better final results by operating a higher amount of material.
Regarding the stock, the PPO2 agent keeps less material than baseline in scenarios with stochastic lead times and has closer levels when considering constant lead times.

\subsection{Learning Curves}
\label{subsec:learning_curve}

\begin{figure*}
  \includegraphics[width=\textwidth]{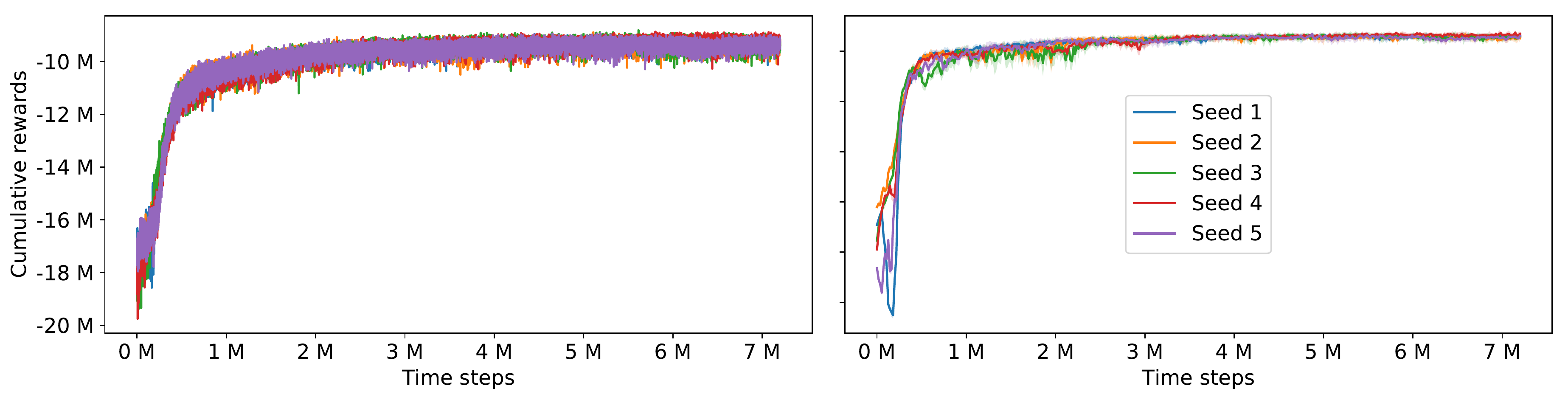}
\caption{Learning curves for scenario \textit{N20}: the left graph shows the actual learning curves regarding the five training runs; the right one shows the mean value of the evaluations carried out during the same training runs.}
\label{fig:learning_curves}       
\end{figure*}

The learning curves of PPO2 agent training, for scenario \textit{N20}, are shown in Figure \ref{fig:learning_curves}.
The vertical axis refers to the total cumulative rewards by episode, and the horizontal axis refers to the number of time steps during training.
The left graph shows the actual learning curve for each one of the five training runs.
As the PPO2 agent learns a stochastic policy, the learning curve is a lower bound of the performance of the algorithm \citep{stable-baselines3}.
So, to better evaluate such a metric, the right graph shows the mean values related to the evaluations carried out during training.
As mentioned in the proposed experimental methodology (Section \ref{subsec:exp_meth}), the model is evaluated on every 50 episodes, or 18 thousand time steps, considering 10 episodes on each evaluation step.
So, the values shown in the right graph refer to the mean of those 10 episodes for each evaluation step.
We can see that, at the beginning of the training, cumulative rewards are between -20 and -16 million, i.e., the total operating costs of the initial solutions are in the order of 16 to 20 million.
After an initial period of exploration, the results start to get better quickly, and the cumulative rewards achieve -10 million after around 1 million time steps of training.
After this point, the improvements are slower and the values tend to converge after 4 million time steps.
As presented, the final solution for this scenario is around -9.3 million.
These curves show that the PPO2 agent was able to learn how to operate the supply chain from a kind of random solution.
They show also that the learning process stabilizes before the end of the training.
Learning curves for other scenarios follow roughly the same pattern and are not presented here for the sake of space.

Concerning the execution time of the algorithm, on average, each training run spent less than 220 minutes.
It is important to note that, after the RL model has been trained, its application has a very short execution time.
Just provide the current state of the supply chain for the model and the neural network outputs will immediately provide the decisions for the next time step.
Thus, even though the training time may have a substantial execution time, its application is fast, which is a considerable advantage in possible real-time decision-making scenarios.

\subsection{Results Summary}
\label{subsec:summary}

The experiments have shown that PPO2 can be a good tool to use in scenarios with stochastic lead times, regardless of whether demands are seasonal or not.
In these scenarios, PPO2 was better than baseline (LP agent) mainly by better meeting the uncertain demands.
These results were achieved by operating a higher amount of material while better attend the stock capacities.
The algorithm can also be useful in scenarios with constant lead times and non-seasonal demands, especially with higher demand uncertainty.
Regarding scenarios with constant lead times and seasonal demands, the PPO2 algorithm and the baseline (LP agent) have achieved similar results.
In such situations, PPO2 would be a more viable option if it is difficult to model the problem or to get accurate values for its parameters.
Considering all scenarios, the greater the uncertainty bigger are the costs of the solutions found by the algorithm, as would be expected.

We have also verified that PPO2 can build stock with seasonality.
The results have shown that the agent starts to build stocks before demands start rising and that higher stock levels are used when the lead times are stochastic.
Finally, the learning curves have shown that PPO2 was able to learn how to operate the supply chain and that the learning process stabilizes before the end of the training.

\subsubsection{Managerial Implications}

In this work, we have addressed the supply chain problem in a production planning approach, i.e., the decisions of the whole supply chain are based on ultimate customer demands, as recommended by \cite{Lee1997} to counteract the bullwhip effect.
There is a single agent that controls all the chain operations as a central decision-maker.
The results of the experiments have shown that, in this context, the PPO2 algorithm can be a good practical choice, especially if the lead times are stochastic, or the demands have higher uncertainty.
In scenarios with lower demand uncertainty (and constant lead times), strategies based on forecast or average values can easily handle the problem, but bigger the uncertainty more difficult to get good results with such type of approach.
The PPO2 algorithm is a policy-based algorithm that approximates the policy using ANNs.
Thus, it can handle the problem without the need to aggregate the state or action values and, thus, can better explore the solution space.
The results have shown that the algorithm can build stocks with seasonality minimizing the bullwhip effect issues.

Another characteristic of the PPO2 algorithm is that the final model (the solution) can be improved by continuing the agent's training.
This can be interesting to adapt the model after a change in the practical scenario, e.g., a new distribution of the demands or lead times, or a modification in some capacity, etc.
Finally, as a model-free Deep RL method, the proposed solution method only needs the simulation of the supply chain.
This can be an advantage in scenarios in which it is difficult to get a precise model of the supply chain operation or to get accurate values for its parameters.

\section{Conclusions}
\label{sec:conclusions}

Decision-making under uncertainty has a strong practical appeal in logistics management due to the inherent complexities of the involved processes.
Artificial Intelligence applications in supply chain planning problems can be a way to improve logistics management and have been increasingly explored in the literature. 
In the present work, we have used a Deep RL approach (PPO2) to solve a production planning and product distribution problem in a multi-echelon supply chain with uncertain seasonal demands and lead times.
We have explored 17 different scenarios in a supply chain with four echelons and two nodes per echelon considering a planning horizon of 360 time steps.
On each time step, the RL agent needs to decide how much raw material to produce in the first echelon nodes and the amount of material to be sent from each node to the nodes of the next echelon (stock levels are indirectly defined).
The goal is to meet uncertain customer demands in the last echelon nodes while minimizing all incurred costs (operation costs, such as production at suppliers, stock, transport, processing; and penalization costs: if demand is not met, or a stock capacity is exceeded).
We have built upon our previous work \citep{AlvesICCL} adding uncertain seasonal demands, stochastic lead times, and manufacturers' capacities. 
The formalization of the problem, an MDP formulation and an NLP model, have been extended to take account of changes in the problem.
To the best of our knowledge, the present and our previous works are the first ones to use Deep RL to handle the problem with a production planning approach in a supply chain with more than two echelons.
Therefore the problem solved has more state and action spaces dimensions, being harder to solve than related works.
Another contribution is that we are the first to solve the problem with stochastic lead times using a Deep RL method, even if we consider related works that handle the problem in an order-based approach.

We have done a robust experimental methodology to verify the quality and suitability of the PPO2 algorithm on the proposed problem.
Firstly, we have conducted a hyperparameter tuning process to choose the best values for the algorithm's hyperparameters. 
Next, we have used such values to solve the problem in different scenarios, considering multiple training runs with different random seeds.
Finally, the results have been evaluated considering 100 episodes for each trained model.
We have compared the achieved results with an LP agent baseline, built from the solution of an LP model, considering forecast demands and average lead times.
PPO2 agent is better than baseline in all scenarios with stochastic lead times (7.3-11.2\%), regardless of whether demands are seasonal or not.
In scenarios with constant lead times, the PPO2 agent is better when uncertain demands are non-seasonal (2.2-4.7\%).
If uncertain demands are seasonal and lead times are constant, PPO2 and LP have roughly the same performance.
Considering the experimental results, PPO2 is a competitive and suitable tool for the addressed problem, and that the greater the uncertainty of the scenario, the greater the viability of this type of approach.
A detailed analysis regarding seasonal stock building and type of costs are also presented and discussed (Sections \ref{subsec:stocks_seasonality} and \ref{subsec:cost_types}).

In real-world OR problems, uncertainties in the parameters of a planning model are very common.
As a model-free approach, Deep RL techniques can be useful in such situations, which could avoid excess capacities and stocks.
Another advantage would be in real-time problems, in which the execution time of a previously trained Deep RL model is very fast.
In future works, we intend to compare PPO2 with other Deep RL algorithms to verify which is the most appropriate RL technique for the proposed problem.
Another possible path is to use stochastic programming approaches to solve the NLP model and compare it with those Deep RL algorithms.



%

\section*{Conflict of interest}

The authors declare that they have no conflict of interest.

\bibliographystyle{spbasic}      
\bibliography{article.bib}   

\end{document}